\pdfoutput=1

\documentclass[11pt]{article}

\usepackage[final]{acl}

\usepackage{times}
\usepackage{latexsym}

\usepackage{amsmath}
\DeclareMathOperator*{\argmax}{arg\,max}

\usepackage{subfigure}

\usepackage{algpseudocode}

\usepackage[T1]{fontenc}

\usepackage[utf8]{inputenc}

\usepackage{booktabs}

\usepackage{microtype}

\usepackage{inconsolata}

\usepackage{multirow}

\usepackage{graphicx}

\newcommand{\increase}[1]{\small{\textcolor{teal}{#1}}}
\newcommand{\worst}[1]{\colorbox{red!20}{#1}}
\newcommand{\best}[1]{\colorbox{green!40}{\textbf{#1}}}
\newcommand{\improved}[1]{\colorbox{orange!40}{#1}}
\newcommand{\decrease}[1]{\small{\textcolor{red}{#1}}}

\title{Evaluating Vision-Language Models for Emotion Recognition}

\author{
\textbf{Sree Bhattacharyya\textsuperscript{1}},
\textbf{James Z. Wang\textsuperscript{1}} \\
\textsuperscript{1}College of Information Sciences and Technology, \\ 
The Pennsylvania State University \\
\texttt{sreeb@psu.edu}, \texttt{jzw11@psu.edu}
}

\begin{document}
\maketitle
\begin{abstract}

Large Vision-Language Models (VLMs) have achieved unprecedented success in several objective multimodal reasoning tasks. However, to further enhance their capabilities of empathetic and effective communication with humans, improving how VLMs process and understand emotions is crucial. Despite significant research attention on improving affective understanding, there is a lack of detailed evaluations of VLMs for emotion-related tasks, which can potentially help inform downstream fine-tuning efforts. In this work, we present the first comprehensive evaluation of VLMs for recognizing \textit{evoked} emotions from images. We create a benchmark for the task of evoked emotion recognition and study the performance of VLMs for this task, from perspectives of correctness and robustness. Through several experiments, we demonstrate important factors that emotion recognition performance depends on, and also characterize the various errors made by VLMs in the process. Finally, we pinpoint potential causes for errors through a human evaluation study. We use our experimental results to inform recommendations for the future of emotion research in the context of VLMs.

\end{abstract}

\section{Introduction}
\label{sec:introduction}

Equipping Artificial Intelligence (AI) systems with the capability to understand emotions is important for sensitive and effective interaction with human users in diverse applications \cite{kolakowska2014emotion, zhao2018predicting, yang2021stimuli, wang2023unlocking}. This has been approached in the past through development of deep architectures, including multimodal and context-aware methods suited for specific downstream applications \cite{lee2019context, mittal2020emoticon, hoang2021context}. The advent of Large Language Models (LLMs), however, has brought about a significant shift in focus. LLMs are now adapted or tuned to achieve what task-specific deep learning models were employed for. As a first step in understanding the inherent capabilities of popular large general-purpose models, before adapting them for specific tasks, LLMs have been evaluated through multi-faceted benchmarking experiments. This ranges from evaluating LLMs in objective \cite{hendrycks2021measuring, lu2022learn} and subjective task settings \cite{ziems2022moral, khandelwal2024moral, fung2024massively}. 

Studies exploring emotions in the context of LLMs span both benchmarking and tuning efforts \cite{xie2024emovit, xenos2024vllms, etesam2024contextual}. Several works focus on evaluating text-only language models for emotional capabilities \cite{liu2024emollms, wang2023emotional} or the use of emotional stimuli to enhance the performance of LLMs in other tasks \cite{li2023large, li2024good}. A few recent works also venture beyond the single modality of text, to approximate the human process of emotion perception more closely. Such works focus primarily on tuning large Vision-language models (VLMs) \cite{xie2024emovit, xenos2024vllms, etesam2024contextual}. However, most of the recent explorations concentrate either on specific datasets and models or directly target resource-intensive instruction tuning without highlighting the specific need to do so. While they present impressive results on overall quantitative metrics, there remains a notable lack of comprehensive and critical evaluation studies to illuminate the precise capabilities, weaknesses, and vulnerabilities of large models when performing emotion recognition in a multimodal setting. 

To address this gap, in this paper, we present an extensive evaluation of popular VLMs for emotion recognition. We analyze their performance from lenses of accuracy and robustness, while also characterizing the causes for errors made by them. We investigate the specific task of evoked emotion recognition, because of (a) its widespread practical relevance in domains such as social interactions \cite{wieser2012reduced, jyoti2016survey, awal2021angrybert}, online e-commerce \cite{sanchez2020opinion}, artistic content creation and recommendation \cite{wang2023unlocking}, etc., and, (b) the non-trivial nature of the task, involving simultaneous multimodal and affective understanding to use implicit affective cues to predict exact, detailed emotions \cite{wang2023unlocking}, which is different from application-oriented tasks where the emotion information is atleast partially present with the model \cite{deng2023socratis, li2024enhancing}. In evaluating VLMs for evoked emotion recognition, we specifically ask the following research questions:  
\begin{itemize}
    \item \textit{RQ1:} How well do VLMs recognize evoked emotions given images and a textual prompt? 
    \item \textit{RQ2:} How robust are the models to minor and major variations in the prompts? 
    \item \textit{RQ3:} What are the types of errors seen in the VLM responses and why do they occur? 
\end{itemize}

We first compile existing image-based emotion datasets to create an \underline{Ev}oked \underline{E}motion benchmark of challenging difficulty, \textsc{EvE}. Using \textsc{EvE}, we evaluate 7 popular VLMs on the task of evoked emotion recognition. Beyond presenting metrics of correctness, in our analysis, we delve deep into additional aspects such as preference exhibited by models towards certain sentiments. We design 8 different settings to study the robustness of models to perturbations in prompts. These include shuffling the order of emotion labels in prompts, open-vocabulary classification, adopting emotional perspectives, and using self-reasoning mechanisms. Finally, we create a formal framework to analyze mistakes made by VLMs and conduct a human study to localize the causes of such mistakes. 

Our key findings show that at the current state, VLMs are inept at predicting emotions evoked by images. We show that VLMs are significantly sensitive to the order in which class labels are presented in the prompts, and perform poorly when no labels are presented. We find that prompting VLMs to adopt an emotional persona has a drastic negative impact on their performance. We also observe that self-reasoning mechanisms help in the case of certain models. This is especially applicable for mechanisms that involve breaking the emotion recognition task down into more tractable sub-components (eg., captioning + reasoning). Finally, through our human study, we elucidate that factors leading to the poor VLM performance pertain not only to the model capabilities but also depend on the data used and task difficulty. We use our findings to further discuss important considerations to improve the emotion perception capability of VLMs. \footnote{We make all code and data available at: \url{https://github.com/sreebhattacharyya/Eve_Benchmark}}.

\section{Related Work}
\label{sec:related_work}
 
Methods studying emotions using LLMs have included using theories grounded in psychology to develop evaluation metrics \cite{wang2023emotional, regan2024can}, generating explanations given suitable image-emotion pairs \cite{deng2023socratis}. Efforts have also been made in the direction of fine-tuning LLMs like LLaMA \cite{touvron2023llama}, BLOOM \cite{workshop2022bloom} to create experts on emotional understanding, through instruction tuning \cite{liu2024emollms}. Training-free enhancement methods have been approached to create emotionally conditioned generations for downstream tasks like image captioning or generating a news headline \cite{li2024enhancing}. 

Few recent works also study emotions with multimodal language models. A recent method proposes visual instruction tuning to improve the performance of open models in evoked emotion prediction \cite{xie2024emovit}, using a resource-intensive method of generating synthetic data and fine-tuning models. Another recent effort evaluates Vision-Language Models (VLMs) for expressed emotion recognition, but includes only a single dataset, and depends on auxiliary models to complete intermediate tasks for the VLMs being evaluated \cite{etesam2024contextual}. Vision-language models have also been employed to generate additional contextual information which is used subsequently for training a Q-Former-based module for expressed emotion prediction \cite{xenos2024vllms}. Despite these promising recent research efforts in the area of emotional understanding with VLMs, to the best of our knowledge, the capabilities of advanced Vision-Language Models in \textit{evoked emotion recognition} have thus far not been comprehensively analyzed. 



\section{Evaluation Data}
\label{sec:benchmark}

We leverage popular, existing, evoked emotion recognition datasets to create \textsc{EvE}, an \underline{Ev}oked \underline{E}motion benchmark for our analysis. This includes EmoSet \cite{yang2023emoset}, FI \cite{you2016building}, Abstract, ArtPhoto \cite{machajdik2010affective} and Emotion6 \cite{peng2015mixed}. The selection of the datasets ensures a diverse range of image types in the benchmark, ranging from images of humans, nature, objects in natural or artistically photographed settings, to images of paintings without any recognizable objects (eg., in Abstract). Emotion6 uses 7 discrete emotion classes, while all other datasets follow Mikel's 8-class emotion model \cite{mikels2005emotional}. The total number of samples in Abstract, ArtPhoto, and Emotion6 are under 2000, and we include the entire datasets for the evaluation. For the larger EmoSet and FI datasets, which contain 118000 and 23184 samples respectively, we downsample them each to contain about 2900 samples, retaining only the most challenging samples, as described below. This is done primarily to limit the time and resource consumption when evaluating closed-source models like GPT. Besides, the large size of these datasets is crucial only when training data-hungry deep learning architectures, and not when evaluating models.

To obtain the downsampled sets, a pre-trained ViT model \cite{dosovitskiy2020vit} is first fine-tuned using the entire EmoSet and FI datasets. This achieves weighted F1 scores of 0.91 and 0.53 respectively. For all predictions by the ViT model, the prediction probability is then obtained. This is used to choose moderate to difficult samples, to create initial candidates for the final evaluation sets. The initial candidates for EmoSet and FI are denoted as \(C_e\) and \(C_f\) respectively. The samples incorrectly classified by the fine-tuned model (most difficult) are automatically included in \(C_e\) and \(C_f\). Then, we consider correctly predicted instances, where the probability of prediction is lower than a certain threshold. This probability threshold is chosen empirically to be 0.8, based on the prediction probability distribution over each dataset. Thus, the candidate sets \(C_e\) and \(C_f\) contain incorrectly classified samples, and samples predicted correctly with probability values less than 0.8. Intuitively, the former group of images represents the most difficult category, while the latter group consists of instances that are of intermediate difficulty. Finally, we subsample randomly from these candidate sets to create EmoSet-Hard and FI-Hard, retaining the original emotion class distributions. We include a more detailed account of the subsampling process in the Appendix (\ref{app:finetune_vit_impl}), including a manual analysis of the higher difficulty level of samples in EmoSet-Hard and FI-Hard.

\section{Experimental Setup}
\label{sec:experiments}

We evaluate open-source models LLaVA (7B, 13B) \cite{liu2024visual}, LLaVA-Next (Vicuna 7B, 13B, Mistral 7B) \cite{liu2024llavanext}, and Qwen-VL \cite{bai2023qwen} along with GPT4-omni \cite{achiam2023gpt}, in a zero-shot manner on the created benchmark. The task precisely requires the models to predict what emotion might be elicited from an individual when they are exposed to the visual stimuli of each image sample in the datasets. We categorize our main experiments into two primary settings: (a) \textit{simple multimodal classification}, where each model is prompted to generate a single-word emotion prediction, and (b) \textit{experiments studying model robustness}, where several minor and major perturbations in the prompts are introduced to study differences in model performances. 

\paragraph{Preliminaries for the task.} For a single iteration of the evaluation process, the inputs are an image \(I\), a prompt \(P\) describing emotion labels (words) for \(k\) discrete emotion classes, \(C = \{c_0, c_1, ... , c_k\}\), where \(C\) represents the set of all emotion labels. Model \(M\), with parameters \(\theta_{M}\), performs the classification operation \(M(\cdot)\) on these inputs, generating a response containing the predicted evoked emotion. The responses are parsed and string-matched with the ground truth class labels, and weighted F1 scores are calculated.

\paragraph{Emotion Properties Analyzed.} We leverage properties of the fine-grained emotion classes in the data to provide a formal framework for our analysis. The fine-grained emotion class labels can be more broadly classified to belong to either positive or negative \underline{sentiment} categories (Refer \ref{app:emotion_properties}). We define "\textit{Sentiment Bias}" using this categorization to help reveal insightful trends in the model performances. We define a model's positive sentiment bias as its exhibited preference towards predicting a true negative sentiment sample to a positive sentiment class, and vice versa. Formally, given model \(M\), for a single image sample, given the ground truth label class \(l\) and model predicted class \(c\), and the sets of positive and negative emotions \(S_P\) and \(S_N\) respectively, we define the positive and negative sentiment bias as:
    \begin{equation}
        p_p = p\,(c \in S_P\, |\, l \in S_N).
    \end{equation}
    \begin{equation}
        p_n = p\,(c \in S_N\, |\, l \in S_P).
    \end{equation}

Using this framework for analysis, we now describe our experiments and key results.

\begin{table*}[t]
\small
    \centering
    \begin{tabular}{cccccc}
        \toprule
         Model & Emotion6 & Abstract & ArtPhoto & FI-Hard & EmoSet-Hard \\
         \midrule
         Qwen-VL & 0.461 & 0.21 & 0.36 & \colorbox{red!20}{0.32} & 0.42 \\
         \midrule
         LLaVA (7B) & \colorbox{red!20}{0.372} & \colorbox{green!20}{\textbf{0.27}} & \colorbox{red!20}{0.22} & 0.42 & \colorbox{red!20}{0.13} \\
         LLaVA (13B) & 0.577 & 0.21 & 0.373 & 0.385 & 0.367 \\
         \midrule
         LLaVA-NEXT (Vicuna 7B) & 0.541 & 0.234 & 0.308 & \colorbox{green!20}{\textbf{0.449}} & 0.26 \\
         LLaVA-NEXT (Mistral 7B) & 0.601 & \colorbox{red!40}{0.079} & 0.364 &  0.374 & 0.401 \\
         LLaVA-NEXT (Vicuna 13B) & 0.593 & 0.162 & 0.350 & 0.368 & 0.341 \\
         \midrule
         GPT4-o & \colorbox{green!40}{\textbf{0.635}} & 0.196 & \colorbox{green!20}{\textbf{0.45}} & 0.42 & \colorbox{green!20}{\textbf{0.503}} \\
         \midrule  
    \end{tabular}
    \caption{F1 scores for Simple Multimodal Classification on \textsc{EvE}. The best and worst-performing models on each dataset are highlighted in green and red colors respectively.}
    \label{tab:exp1_results}
\end{table*}

\begin{table}[t]
\small
    \centering
    \begin{tabular}{ccc}
    \toprule
         Model Family & Positive Emotions & Negative Emotions \\
         \midrule
         Qwen-VL & 0.33 & \textbf{0.35} \\
         LLaVA & \textbf{0.30} & 0.29 \\
         LLaVA-Next & \textbf{0.34} & 0.32 \\
         GPT4-o & 0.38 & \textbf{0.48} \\
         \bottomrule
    \end{tabular}
    \caption{Average F1 scores for samples belonging to the broader positive and negative sentiment categories.}
    \label{tab:sentiment_avg_f1}
\end{table}

\begin{figure}[t]
\centering
    \includegraphics[width=\columnwidth]{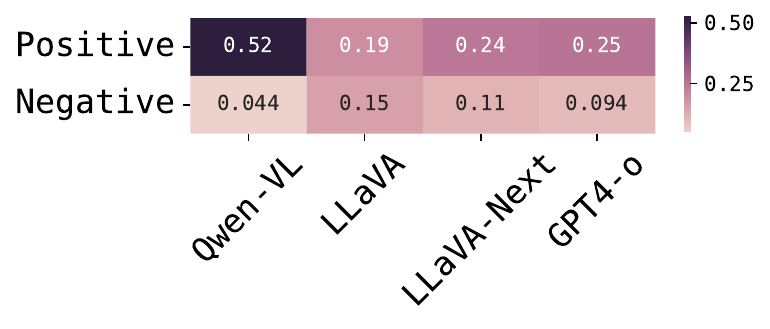}
    \caption{Positive and Negative Bias demonstrated by the models in simple multimodal classification. Results are averaged across datasets and model sizes.}
    \label{fig:sentiment_bias_simple}
\end{figure}

\section{Simple Classification: How well do VLMs perform emotion recognition? [RQ1]}
\label{sec:exp_rq1}

Our first and simplest evaluation scheme, \(M_{\text{c}}(\cdot)\), denoting simple classification, involves prompting the models to choose a single emotion word from the list of labels provided in the prompt. Formally, each model generates:
\begin{equation}
O_{c} = M_{c}\,(I, P_{c}, C;\, \theta_{M}) = c_{j}    
\end{equation}
where \(j \in \{0, ..., k\}\).

From the results described in Table \ref{tab:exp1_results}, we note that the performance is determined not only by the model used but also by the content of the dataset on which it is evaluated. GPT4-o consistently outperforms most open models and even rivals the performance of fine-tuned models on certain datasets \cite{xie2024emovit, xu2022mdan} like Emotion6. 
Despite that, along with all other models, it falls short on the Abstract dataset, which contains images of abstract paintings without any human figures or objects. Further, open-source models LLaVA and LLaVA-Next outperform GPT4-o specifically on the FI dataset. 
Although some models perform comparably to fine-tuned or trained architectures on some datasets, overall, the zero-shot performance of VLMs in emotion recognition still largely lags behind models created specifically for this task. 

We also look at the broader sentiment categories that the data samples belong to \footnote{For all fine-grained emotion class-related analysis, we include only the data subsets following the 8-class emotion model, for uniformity and ease of classification into broader sentiment and arousal categories.}. In Table \ref{tab:sentiment_avg_f1}, we report the average F1 scores achieved by each model on each overarching sentiment category. For all models other than GPT4-o, the difference in performance on positive and negative sentiments is marginal. Models from the the LLaVA family perform slightly better on positive emotions, while Qwen-VL and GPT4-o are better on negative emotions. GPT4-o, despite showing the largest different between the two sentiment categories, has the highest individual F1 score for both sentiments. 

Diving deeper, we calculate the sentiment bias exhibited by the models (Fig. \ref{fig:sentiment_bias_simple}). We observe that models prefer positive sentiments over negative sentiments with a higher probability. This shows that when not fine-tuned specifically for emotion-related tasks, and provided with emotion class labels, all of the models naturally exhibit a higher tendency to generate predictions of positive sentiments. 

In our subsequent experiments, we aim to understand whether the model performance, along with the exhibited biases, is dependent on the specific format of prompts and responses. 

\begin{figure*}
    \centering
    \subfigure[]{
    \includegraphics[width=0.48\textwidth]{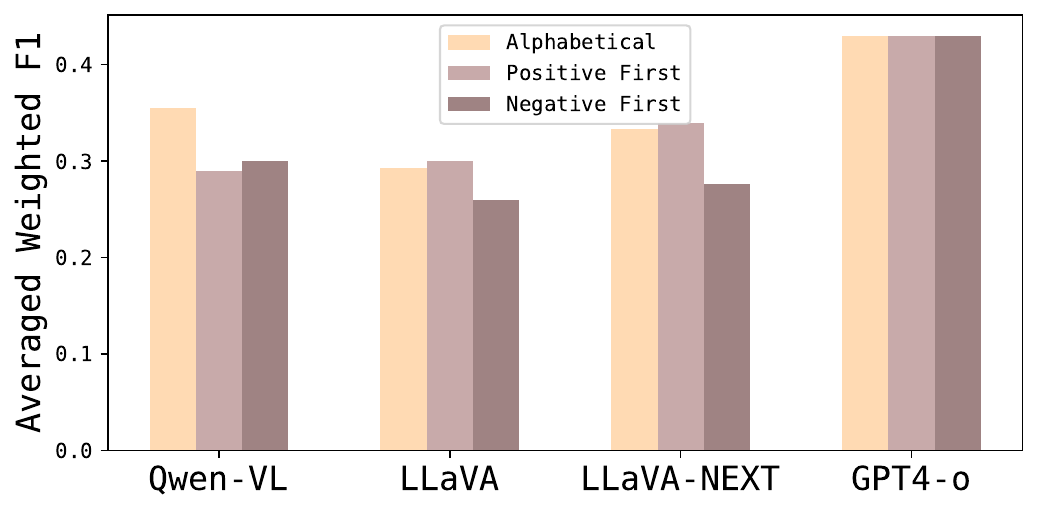}
    \label{fig:shuffled_emotion_order_average}}
    \hfill
    \subfigure[]{
    \includegraphics[width=0.48\textwidth]{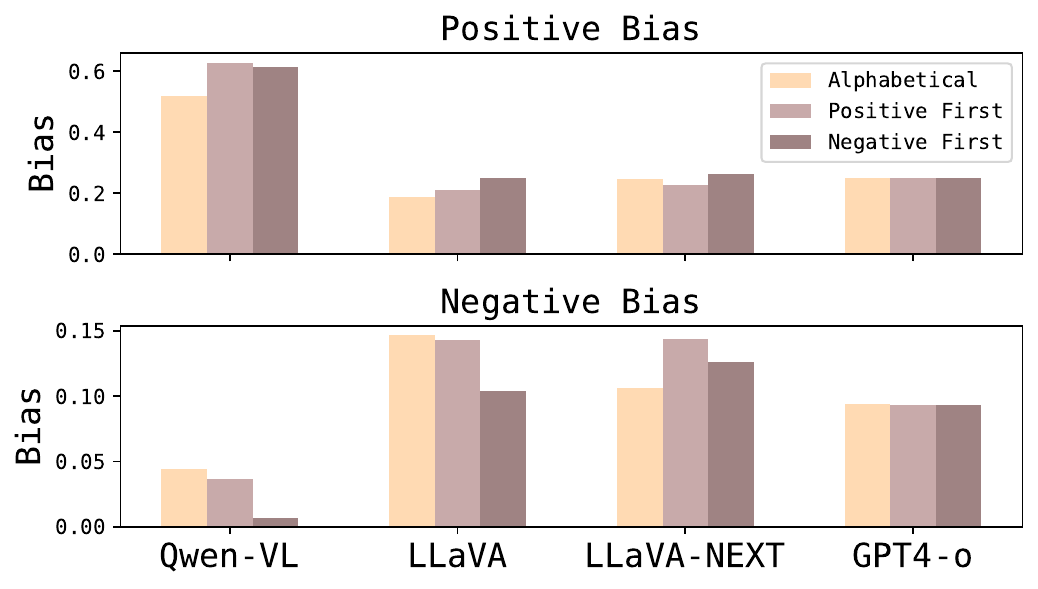}
    \label{fig:sentiment_bias_shuffled}}
    \caption{(a): The weighted F1 score for each model, averaged across datasets. The different bars represent the orders in which emotion class labels are included in the prompt. (b): The positive and negative sentiment bias, for each model, with different shuffled orders of emotion classes in the prompts.}
\end{figure*}

\section{Robustness: How robust are VLMs to changes in emotion-related prompts? [RQ2]}
\label{sec:exp_rq2}

We experiment with four types of changes to study the sensitivity of models: (a) shuffling the order of class labels in the prompts, (b) providing no class labels, (c) adopting an optimistic or a pessimistic persona, and (d) using three different self-reasoning mechanisms. This is mainly to understand whether the models get easily affected by the order or absence of class labels, gauge whether assuming a differing perspective improves or deteriorates the model performance and understand the effect of reasoning strategies which have been shown to be helpful in wide-ranging tasks \cite{wei2022chain, li2024enhancing}. 

\subsection{Variation 1: Shuffled Emotion Order}
\label{sec:shuffled_emotion}

The experiments in the previous section present the emotion class labels in alphabetical order in the prompts. In this section, we explore whether listing any one category of emotions (positive or negative) first, within the prompt, has an impact on the emotion recognition capability of the models. For example, with Mikel's 8-class model \cite{mikels2005emotional}, presenting positive emotions first in the prompt would mean adhering to the following order: \textbf{amusement, awe, contentment, excitement} and \textbf{anger, disgust, fear, sadness}.

Fig. \ref{fig:shuffled_emotion_order_average} reports the weighted F1 scores for all models, averaged across datasets and model sizes. For all models, including the \textit{negative emotion labels first leads to lower performance}. For LLaVA and LLaVA-Next, prompts that have positive emotion labels first show a slight performance improvement. Listing negative emotions first, on the other hand, leads to lower performance for all models, other than GPT4-o, which remains unaffected. We further unveil the precise impact of the shuffled order of emotion labels on sentiment bias. As shown in Fig. \ref{fig:sentiment_bias_shuffled}, for all models other than GPT4-o, positive sentiment bias generally increases when \textit{either emotion class} is presented first. Conversely, negative bias generally decreases with both kinds of shuffled order of emotions, except for LLaVA-Next. Thus, overall, most open-source models deteriorate when negative emotions are presented first, while their positive bias is increased when emotions are grouped according to sentiment categories. 

\begin{figure}[t]
    \centering
    \includegraphics[width=\columnwidth]{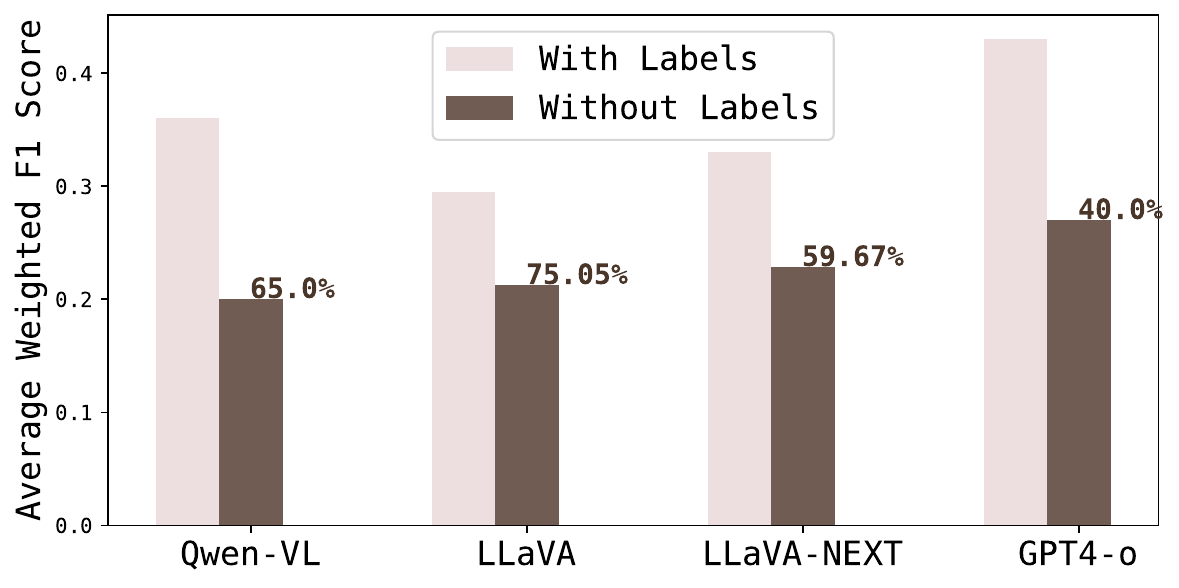}
    \caption{The weighted F1 score with and without precise target labels in the prompts. The numbers in brown represent the percentage of fine-grained predictions made.}
    \label{fig:no_label_f1}
\end{figure}

\begin{figure}[t]
    \centering
    \includegraphics[width=0.9\columnwidth]{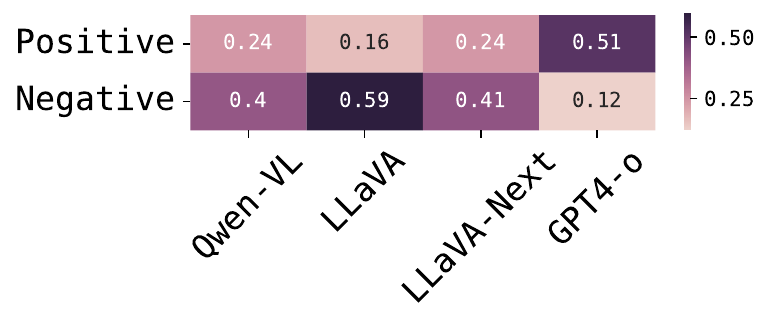}
    \caption{Sentiment bias for responses generated without explicit target labels in the prompts.}
    \label{fig:no_label_sentiment_bias}
\end{figure}

\subsection{Variation 2: Providing No Target Labels}
\label{sec:no_label}

For experiments in this section, we provide no explicit emotion class labels in the prompt to choose from. The models are free to respond using a single emotion word that does not necessarily belong to the datasets' label set. We use semantic similarity scores generated using SBERT \cite{reimers2019sentencebert} to assign the predictions to the class with the most semantically similar label. As our task involves fine-grained emotion recognition, we further consider whether the free-form predictions by the models are specific enough. Using the original class labels from the datasets, \(C = \{c_1, c_2, ...., c_k\}\), we calculate: 
\begin{equation}
\label{eq:maximum_similarity_distinct}
    sim_{\text{max}} = \max_{i,j}\, (sim\,(c_i, c_j)),\, i \neq j 
\end{equation}
Intuitively, it denotes the maximum possible similarity between two \textit{distinct, fine-grained} emotion classes. Thus, for each free-form prediction to be sufficiently specific, its similarity to the correctly assigned label class should be greater than the maximum similarity between two distinct classes. Given the set of all open-vocabulary model predictions \(O\), and the ground truth labels \(L\), we calculate the frequency with which each model makes adequately fine-grained predictions as follows: 
\begin{multline}
    p\,(sim(o_i, l_i) > sim_{\text{max}}\, |\, E) \, \forall\, o_i \in O, l_i \in L        
\end{multline}
where \(E\) denotes the event of \(o_i\) being assigned to class \(l_i\).

Fig. \ref{fig:no_label_f1} shows the F1 scores for each model, across datasets, with the frequency of fine-grained predictions depicted through the numbers above the bars. All models fare significantly better when provided with labels in the prompts than when open-vocabulary prediction is required. Note that this is true even when the final classification is done using only maximum semantic similarity, which is a more relaxed criteria than requiring an exact string  match with the provided labels. Further, LLaVA on average makes fine-grained predictions more often than all other models. GPT4-o uses specific emotion words the least often, implying that to make it suitable for use in fine-grained emotion prediction tasks, the inclusion of target labels is indispensable. 
Additionally, we compute the sentiment bias scores (Fig. \ref{fig:no_label_sentiment_bias}), and find that the earlier trend is reversed for all models other than GPT4-o, when compared to classification with explicit target labels in the prompts (Fig. \ref{fig:sentiment_bias_simple}). This also shows, that when the predictions are not anchored using predefined class labels in the prompts, most models have a naturally higher likelihood of choosing negative emotion words over positive emotions.

\begin{figure*}
    \centering
    \subfigure[]{
        \includegraphics[width=0.32\textwidth]{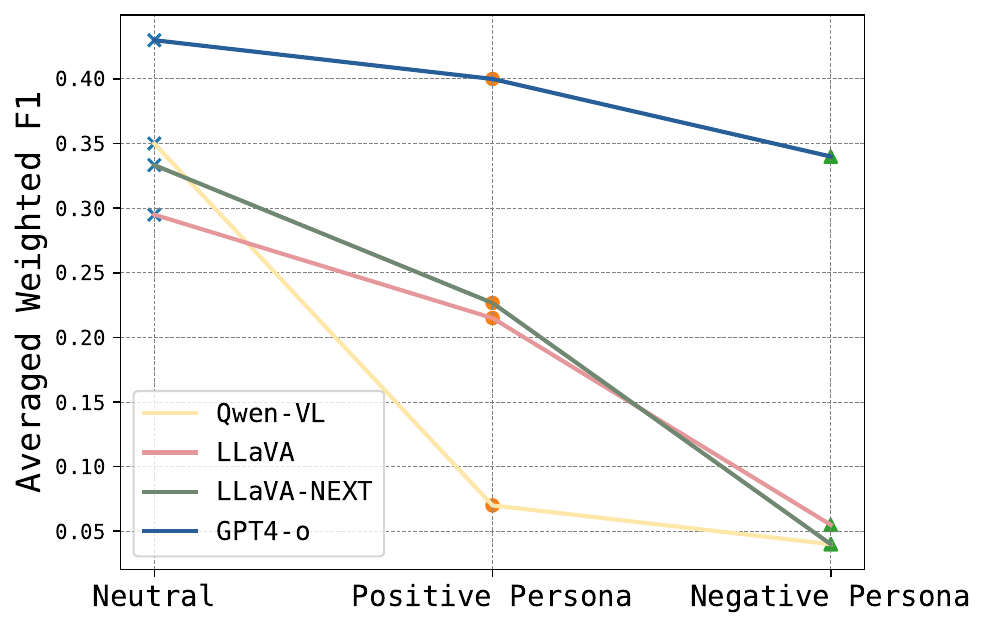}
        \label{fig:persona_f1}}
    \hfill
    \subfigure[]{
        \includegraphics[width=0.32\textwidth]
{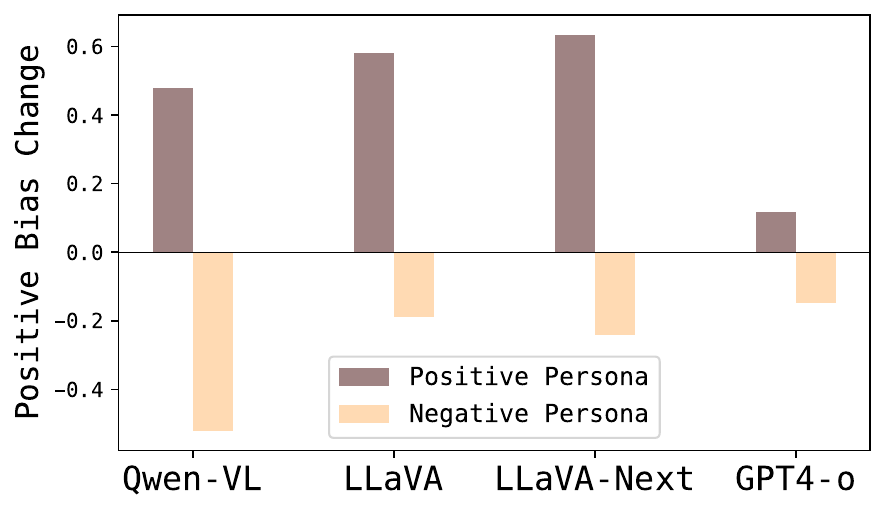}
        \label{fig:persona_positive_bias}}
    \hfill
    \subfigure[]{
        \includegraphics[width=0.31\textwidth]{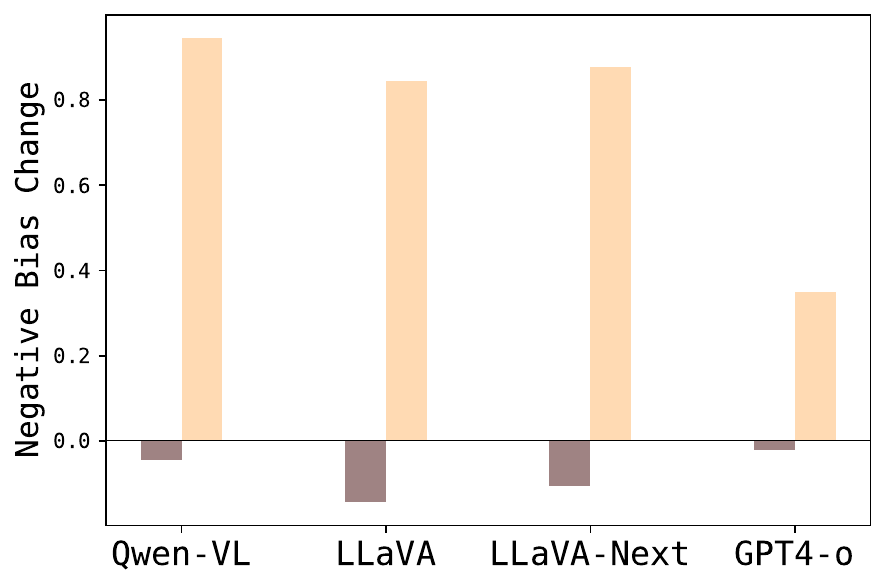}
        \label{fig:persona_negative_bias}}
    \caption{Fig. (a): Weighted F1 score for each model, averaged across all datasets considered. The score drops sharply when the models assume any sentimental persona.  Fig. (b): Change in \textit{Positive Bias} when assuming any persona. Positive Bias is increased and decreased significantly by Positive or Negative Persona. Fig. (c): Change in \textit{Negative Bias} when assuming any persona. Negative Bias is sharply increased when assuming a negative persona but reduced only marginally by positive persona.}
\end{figure*}

\subsection{Variation 3: Adding a Persona}
\label{sec:persona}

Approaching robustness from another angle, we explore whether urging the models to adopt a sentiment-related perspective (positive or negative) holds any influence. Specifically, we study whether adding an optimistic persona biases the model to choosing positive emotions more frequently, and vice versa, besides affecting the overall performance. We plot the average F1 score for each model, under different assumed personas in Fig. \ref{fig:persona_f1}. 

All models perform poorly when adopting either a positive or negative persona. The degradation in performance is the most stark for Qwen-VL, and the least for GPT-4o. Across all the models and datasets, the performance drop when adopting a negative persona is significantly more than when adopting a positive persona. We show changes in the sentiment bias to be a primary reason for the poorer performance, as demonstrated through Fig. \ref{fig:persona_positive_bias} and Fig. \ref{fig:persona_negative_bias}. It can be noted from Fig. \ref{fig:persona_positive_bias}, that adopting a positive persona sharply increases the positive bias, which, on the other hand, is diminished by using a negative persona. 

Similarly, as seen in Fig. \ref{fig:persona_negative_bias}, negative bias increases sharply when adopting a negative persona, leading to models classifying nearly all samples to negative emotion classes (most frequently "sadness"). In contrast to the change in positive bias, negative bias is only marginally reduced when using a positive persona. Thus, all models show extreme vulnerability to the inclusion of a sentimental perspective. This could potentially make models susceptible to exploitation, for inducing severe bias in emotion-related tasks. 

\subsection{Variation 4: Reasoning-based Prompting Mechanisms}
\label{sec:reasoning}

Adapting prompting methods like Chain-of-Thought \cite{wei2022chain}, we explore whether prompting the models to self-reason with their generation impacts the performance. Specifically, we use three different evaluation mechanisms. In the first mechanism, the model generates an explanation for its emotion prediction simultaneously. The second mechanism uses three steps of contextual reasoning prior to prediction. The first two steps involve attending to the foreground and background objects in the images to predict emotions evoked by them individually. The third step requires reasoning about whether these two emotions are compatible, to decide the final prediction. Our last mechanism involves captioning the provided image, followed by reasoning using the caption to predict evoked emotion. 

\begin{figure}
    \centering
    \includegraphics[width=\linewidth]{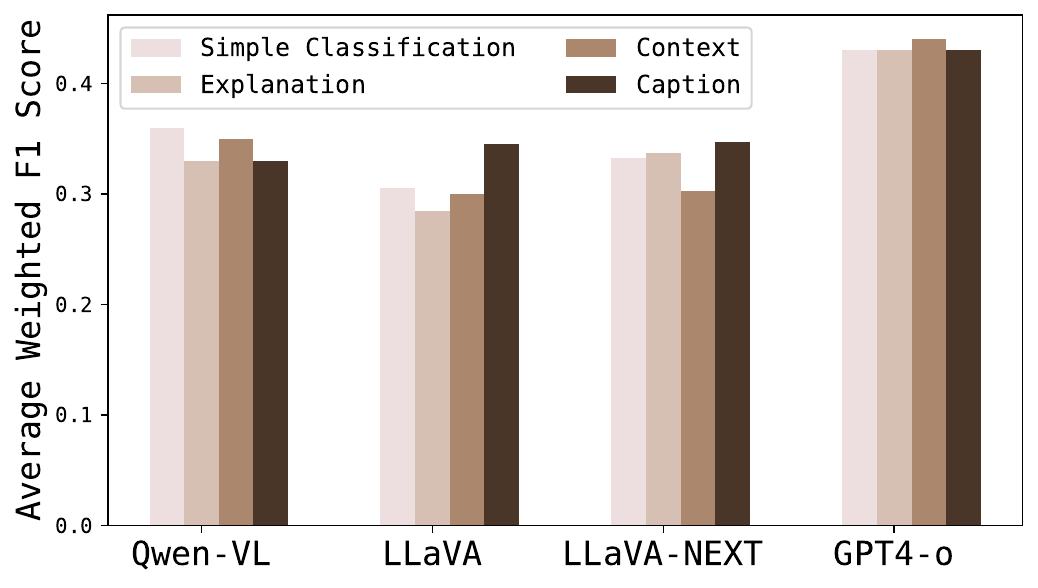}
    \caption{Weighted F1 scores, averaged across all datasets considered, for different prompting mechanisms explored.}
    \label{fig:reasoning_f1}
\end{figure}

The aggregated F1 scores are presented in Fig. \ref{fig:reasoning_f1}. Contextual reasoning helps only GPT4-o among all models, highlighting the inability of most other models to capture relevant background context from images accurately. By analyzing specific responses, we also observe that LLaVA and LLaVA-Next struggle with the multi-step response format required for contextual reasoning. Captioning-based reasoning shows relatively higher gains with LLaVA and LLaVA-Next. This further underscores that these models underperform when reasoning over multiple modalities (image and text) simultaneously, compared to when reasoning only over text (captions of images). Overall, the models remain relatively robust to variations in the prompting mechanism, and only show slight improvements in some cases. 

From the results of our robustness experiments, it can be concluded that most models show significant variance with respect to prompt perturbations. However, the extent of such variance is largely determined by the type of perturbation. Designing models that are robust to such variations is thus an important area for further inquiry.

\section{Analyzing Mistakes by Models [RQ3]}
\label{sec:exp_rq3}

\begin{figure*}[t]
    \centering
    \includegraphics[width=\textwidth]{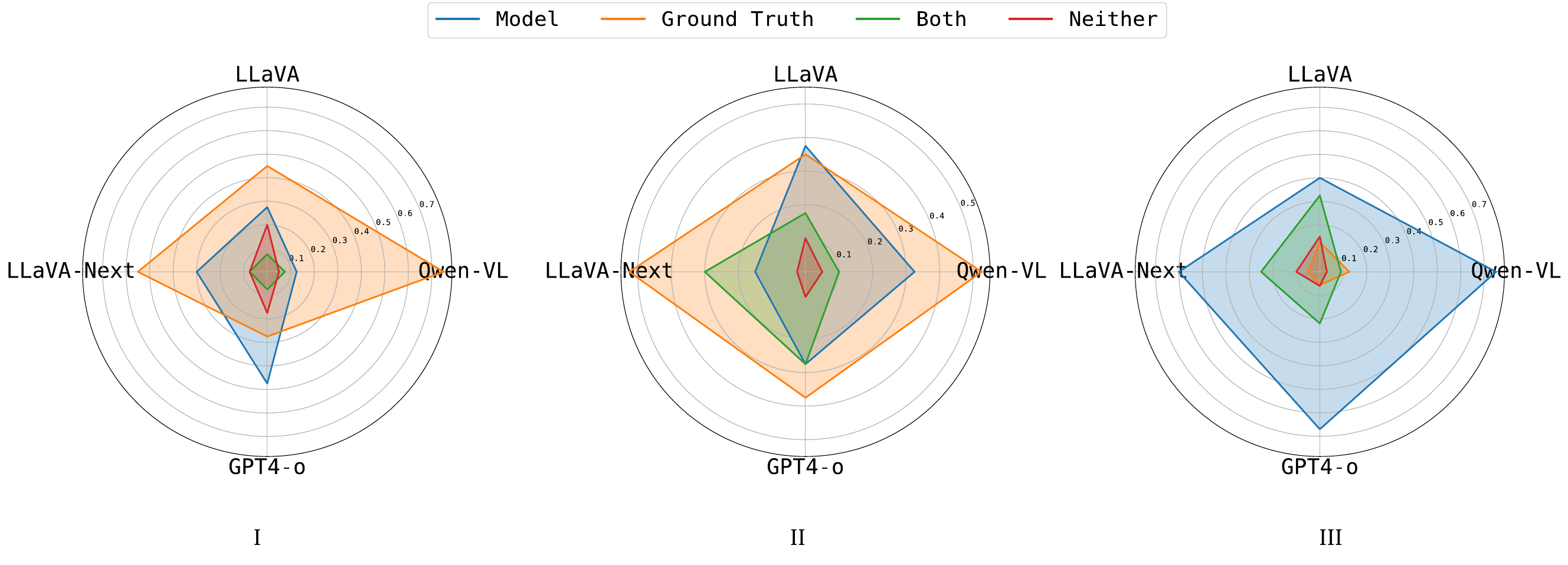}
    \caption{Human agreement with model predictions (blue), ground truth from dataset (orange), both labels (green), and neither label (red) for different models and error categories (from Left to Right: Error Category I, II, III).}
    \label{fig:human_agreement}
\end{figure*}

To characterize errors made by each model, we define three types of errors, proceeding from broad (or blatant) to fine-grained (subtle) mistakes: 
\begin{itemize}
    \item \textit{Error Category (EC) I - Incorrect Sentiment}: The case where the ground truth and predicted label belong to different sentiment categories (eg., "sadness" and "amusement"). 
    \item \textit{Error Category (EC) II - Correct Sentiment, Incorrect Arousal}: The case where the ground truth and predicted label belong to the same sentiment but the different arousal or intensity category (eg., "sadness" and "anger"). Arousal, in the dimensional Valence-Arousal-Dominance (VAD) model of emotions, refers to the agitation level of a person, or the intensity of the emotion felt.
    \item \textit{Error Category (EC) III - Correct Sentiment, Correct Arousal, Incorrect Prediction}: The case where the ground truth and predicted label belong to the same sentiment and arousal/intensity category, but are not the same fine-grained class (eg., "fear" and "anger"). 
\end{itemize}

We hypothesize that blatant errors in EC I can be attributed to the model's inability to reason about affect. However, the more nuanced errors (II and III) could be caused by subjective interpretation of closely related, distinct emotions. We conduct a manual evaluation study to explore this further, annotating about 500 error samples. Each annotation denotes whether a human rater agrees more with the model-predicted emotion label, with the original ground truth label from the dataset, with both emotion labels, or with neither. 

We plot the human agreement percentage in Fig. \ref{fig:human_agreement}. The plot for EC I shows that errors in this category are indeed genuine errors by the models, as human annotations consistently agree more often with dataset ground truth. 

For EC II, although agreement with ground truth still dominates, there is a significant increase in agreement with both model predictions and ground truth, proving that some of the errors in this more fine-grained category can be attributed to the subjectivity of emotion perception. 

Finally, for EC III, most of the model predictions, that do not match with dataset labels at a fine-grained level, may not be entirely incorrect, since they are preferred more often than the dataset ground truth. We also observe specific examples where the dataset ground truth incorrectly reflects the \textit{expressed} emotion, while the model predictions accurately capture the \textit{evoked} emotion. The so-called errors in EC III can thus be attributed to noisy ground truth from the datasets, rather than the capability of VLMs. This unveils the issue of unreliable ground truth labels in existing emotion datasets. It can also be noted that for all the error categories, the proportion of human annotations agreeing with neither the ground truth nor model prediction (labeled as "Neither") remains relatively small and constant. This reflects that either of the two emotion labels (ground truth or model predicted) or both were found to be plausible in most cases, showing that human preference of either category (as measured by agreement of annotations) is clear and trustworthy. We additionally show in the Appendix (\ref{app:humeval}), that the Abstract data subset, on which models perform most poorly, is the most reliable dataset in terms of human agreement.

\section{Discussion and Conclusion}
\label{sec:discussion}

We arrive at answers to our initial research questions through all of our experiments: 

\begin{itemize}
    \item \textit{RQ1}: VLMs are not adept at zero-shot multimodal emotion recognition, and often exhibit significant biases towards certain emotions.
    \item \textit{RQ2}: VLMs are sensitive to prompt changes. The performance depends largely on the way target labels are presented, the format of prompting and response, and whether VLMs adopt a sentimental perspective. 
    \item \textit{RQ3}: VLMs make a combination of broad and fine-grained errors. Many deviations from a dataset's ground truth can also be attributed to ambiguous or unreliable original labels. This is especially applicable for the most fine-grained errors. 
\end{itemize}

The need for improvement in model capabilities could be approached through a deeper investigation of the internal model representations, the methods used for aligning models to the tuning data, etc. However, such interventions would require for the instruction-tuning or fine-tuning data to be noise-free. To make the datasets reliable, while accommodating the inherent subjectivity of the task, datasets could be created with explanations for annotations, emotion distributions or multiple labels instead of discrete single class labels. Further, the research community could benefit from availing detailed information on datasets, such as, the test-retest reliability data \cite{kim2018development}, duration of exposure to emotion stimuli for each subject \cite{lu2017investigation}, etc. Further, there remains a strong need to distinguish between \textit{evoked} and \textit{expressed} emotions. Many current datasets are curated by querying images online using emotional keywords \cite{yang2023emoset}, which is susceptible to collecting images merely related to the keyword, and not necessarily evoking that exact emotion. 

Through our experiments, human study and analysis, we hope to have highlighted that all aspects of VLMs' emotion recognition pipeline, specifically the data used and modeling, are in need of critical analysis and measures for improvements. Through this work, we also hope to inspire broader evaluation and benchmarking efforts to improve emotional reasoning in VLMs, extending to complementary areas of emotion understanding and generation, to help achieve the broader goal of making AI systems more empathetic, safe and useful.

\section{Limitations}
\label{sec:limitation}

Although the current state of the study aims to be the first comprehensive evaluation of VLMs for evoked emotion recognition, there remains scope to include more models. With greater resources, there opens up the possibility of evaluating entire datasets and comparing the same with the model performances on the harder subsets included in our benchmark. 

The current evaluation also includes only few-shot performances of the models, while the opportunity to fine-tune smaller models on the same datasets, particularly the difficult data subsets, remains open. 

Further, the datasets currently included are shown to have ambiguous instances, which stem both from innate subjectivity of emotions and noise. Although we discuss useful measures to reduce make datasets more reliable, the possibility of ambiguous interpretations of emotions is a major challenge in affective computing. It continues to be an active area of research. 

As most of the images are sourced from the internet, we also acknowledge the possibility of some of the images being included in the training data of the models evaluated. However, for a closed model like GPT4-o, it is not possible to verify the same.

The current benchmark and evaluation also address the specific task of evoked emotion recognition and could be extended to include other tasks in emotion recognition, as well as generation, to constitute a comprehensive benchmark for emotional understanding. 

\section{Ethical Considerations}
\label{ethics}

We depend on existing emotion datasets to create our benchmark. We acknowledge that the possibility of offensive images being present in the datasets cannot be ruled out. Although we manually analyze several instances from the datasets, we do not manually check the precise visual content in all of the images. Besides, though the datasets used do not contain any private identifiable information, a large number of images include humans, revealing their faces and gestures. We implore against the misuse of that information and will ensure dissemination of the dataset only for verifiably legitimate and valid purposes of research. As some of the datasets were also created many years ago, it is possible that they may not satisfy the required bar of ethical review in place at present. Ensuring that they do comply with the required standards of reproducibility and reliability can in itself be an important area of research. Finally, we only evaluate how well the models mimic trends it has learned through the multimodal data used for training, and do not claim that they possess any real, human-like, "understanding" of emotions. 

In a larger perspective, our research aims to help create emotionally sensitive VLMs. We acknowledge that depending on the deployment of VLMs, emotional information could potentially be used for manipulating human behavior, such as using positive emotions to advertise products. Although the end result of such deployment is largely determined by the executive forces controlling the use of large models, we advocate strongly responsible usage of our research, and similar research endeavors. We emphasize the need for a thorough risk analysis prior to practically applying emotionally-equipped large language or vision-language models. 

\bibliography{acl_latex}
\appendix
\newpage
\section{Sampling for Benchmark}

In this section, we present additional details about the process adopted for creating the FI-Hard and EmoSet-Hard data subsets. This includes details of implementation, followed by examples of the varying difficulty levels targeted to be included through the benchmark generation process. 

\subsection{Subsampling Data based on Fine-tuning}
\label{app:finetune_vit_impl}

We first present specific implementation details of the fine-tuning process. We use the base size of ViT \cite{dosovitskiy2020vit}, pretrained on ImageNet-21K \cite{ridnik2021imagenet} and fine-tuned on ImageNet 2012 \cite{russakovsky2015imagenet}. For fine-tuning on EmoSet, we add 3 linear layers, each followed by dropout layers (p=0.2) and a non-linearity of ReLU \cite{nair2010rectified}. Training is carried out for 30 epochs, creating an 80:20 split into training and holdout sets. As the number of samples in FI is significantly smaller, we use the model pre-trained on EmoSet as the starting point for fine-tuning on FI. Making the final classification layer trainable, we update the weights of the pretrained model, based on the FI dataset. The training for FI follows a similar 80:20 split of training and unseen validation data, and is carried out for 30 epochs. In both cases, the models are optimized with Stochastic Gradient Descent, using an initial learning rate of 0.05, momentum of 0.9, and weight decay set to 0.00005. Along with SGD, Cosine Annealing scheduler is used. The objective is simply minimizing the multi-class Cross-Entropy Loss. The models are fine-tuned on single A40 GPUs with 4 cores. The total time taken for fine-tuning EmoSet and FI was around 5 hours and 3 hours respectively. 

\begin{table}[t]
\small
    \centering
    \resizebox{\linewidth}{!}{
    \begin{tabular}{ccccc}
    \toprule
         EmoSet-Hard & FI-Hard & Abstract & ArtPhoto & Emotion6 \\
         \midrule
         2998 & 2976 & 250 & 805 & 1980 \\
         \bottomrule
    \end{tabular}}
    \caption{The final number of images in each of datasets used in our benchmark and evaluation experiments.}
    \label{tab:data_stats}
\end{table}

\begin{figure*}
    \centerline{
    \includegraphics[width=\textwidth]{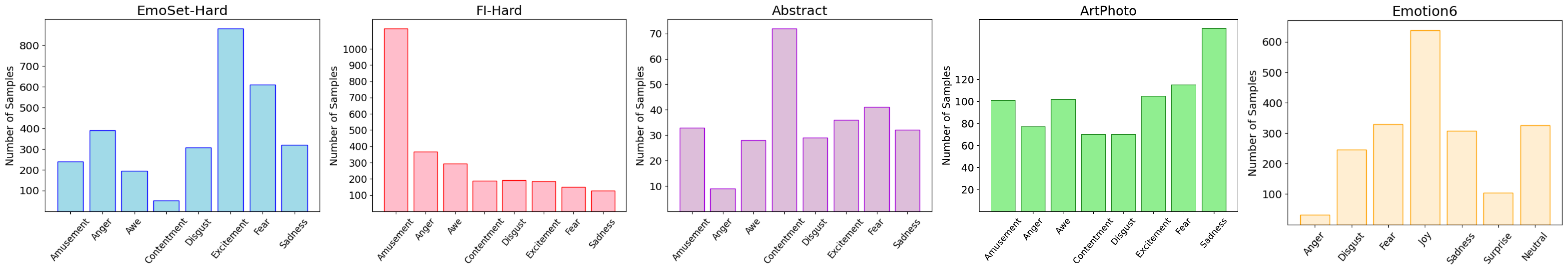}
    }
    \caption{The distribution of different emotion classes in the final evaluation sets considered. The numbers of samples in different emotion classes, in EmoSet-Hard and FI-Hard are proportional to the original class distribution in the candidate sets they are obtained from by subsampling.}
    \label{fig:data_class_dist}
\end{figure*}

\begin{figure*}
    \centerline{
    \includegraphics[width=0.8\textwidth]{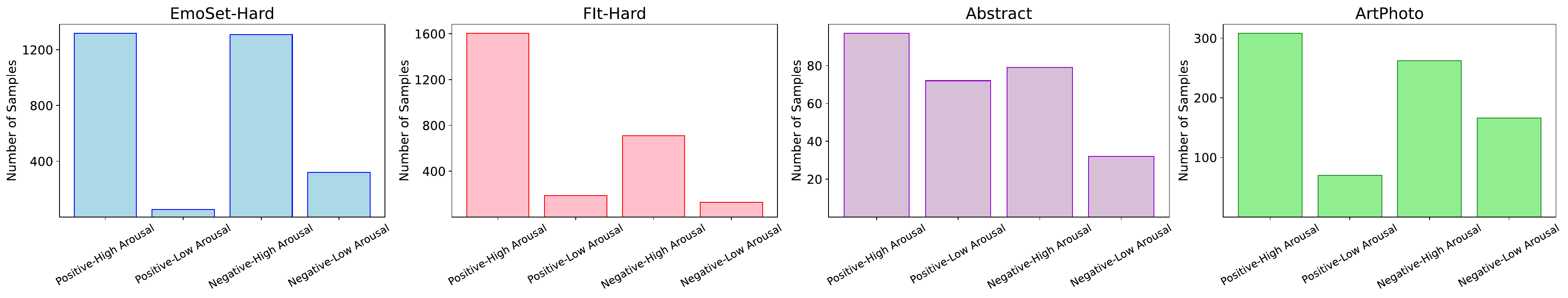}
    }
    \caption{The distribution of different emotion classes in the final evaluation sets considered, grouped according to the broader Sentiment and Arousal categories. The grouping is shown only for the datasets considered in fine-grained class-specific analysis.}
\end{figure*}

Once the model is fine-tuned on the entire EmoSet and FI datasets, as described in the main body, the prediction probabilities are used to further filter out the most obvious or easy samples. The probability values for the correctly classified samples in EmoSet are in the range \([0.31, 1.0]\), and for FI are within in \([0.32, 1.0]\), with most probability values lying above 0.9. We choose the threshold of 0.8 for both EmoSet and FI, keeping the value close to the average of the range, but slightly higher, to account for the higher frequency of probability values greater than 0.9. Thus, the final data subsets contain samples that are either incorrectly predicted by the fine-tuned model, or are predicted correctly with probability less than 0.8. Intuitively, it includes examples that are harder to classify, contain less obvious expressions of emotion, or can potentially belong to multiple emotion classes. The final numbers of samples in each data subset is described in Table \ref{tab:data_stats}. In the final subsamples, we also retain the original emotion class distribution of each dataset, as represented through Fig. \ref{fig:data_class_dist}. As we do not train or fine-tune any models, the varied class distribution is not detrimental to our analysis. Further, to account for the unequal class distribution, we report the weighted F1 scores for all analysis.

\begin{figure*}
    \centerline{
    \includegraphics[scale=0.55]{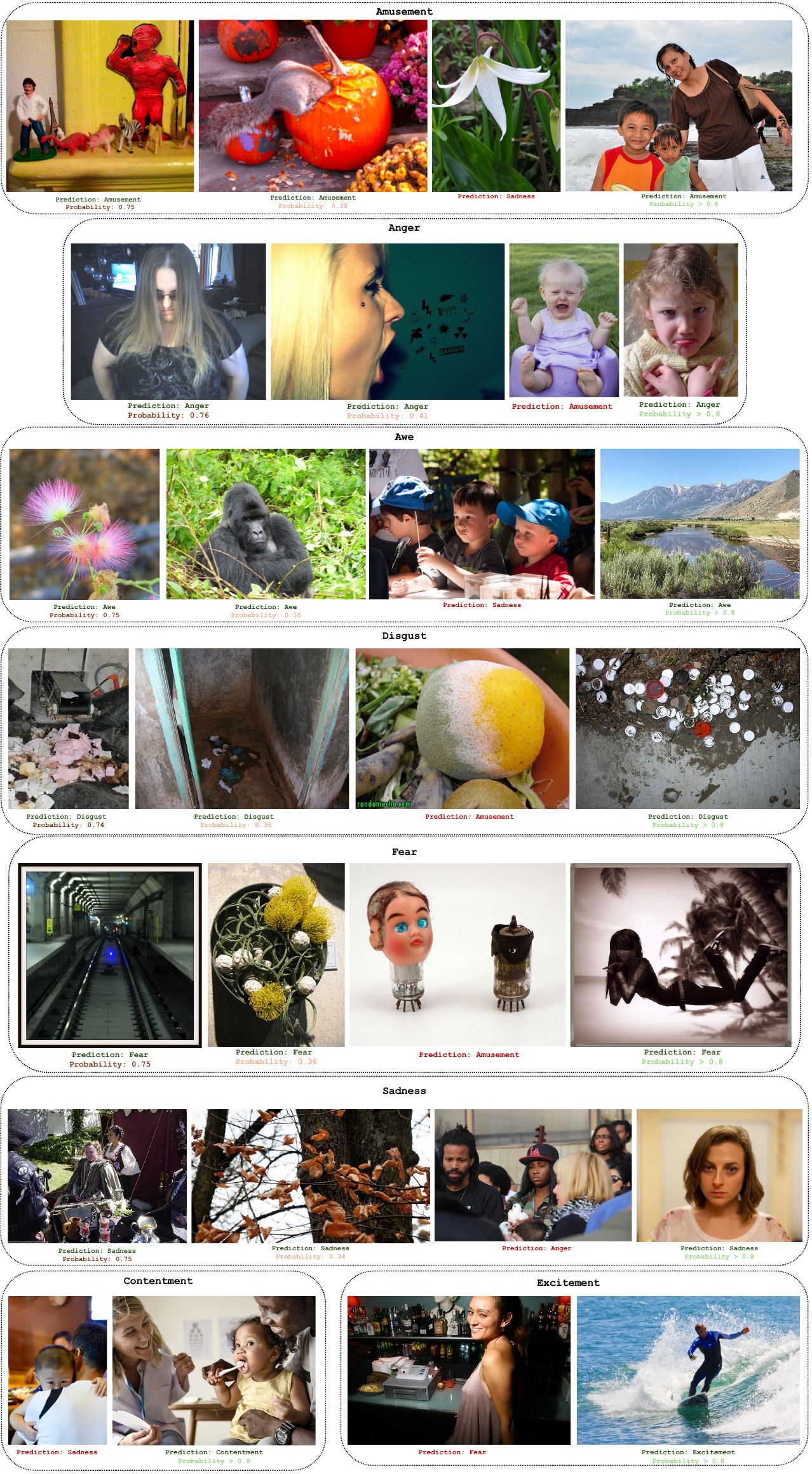}}
    \caption{Examples from the created \textbf{EmoSet-Hard} dataset. For Contentment and Excitement, no instances are found that are predicted correctly with a probability less than 0.8. For all other categories, the two leftmost examples describe instances that are correctly predicted, but with a probability less than 0.8. The next example shows an image predicted incorrectly. Finally, the rightmost example for all categories show the correctly predicted samples, which have probability of prediction higher than 0.8.}
    \label{fig:benchmark_example_emoset}
\end{figure*}

\begin{figure*}
    \centerline{
    \includegraphics[scale=0.8]{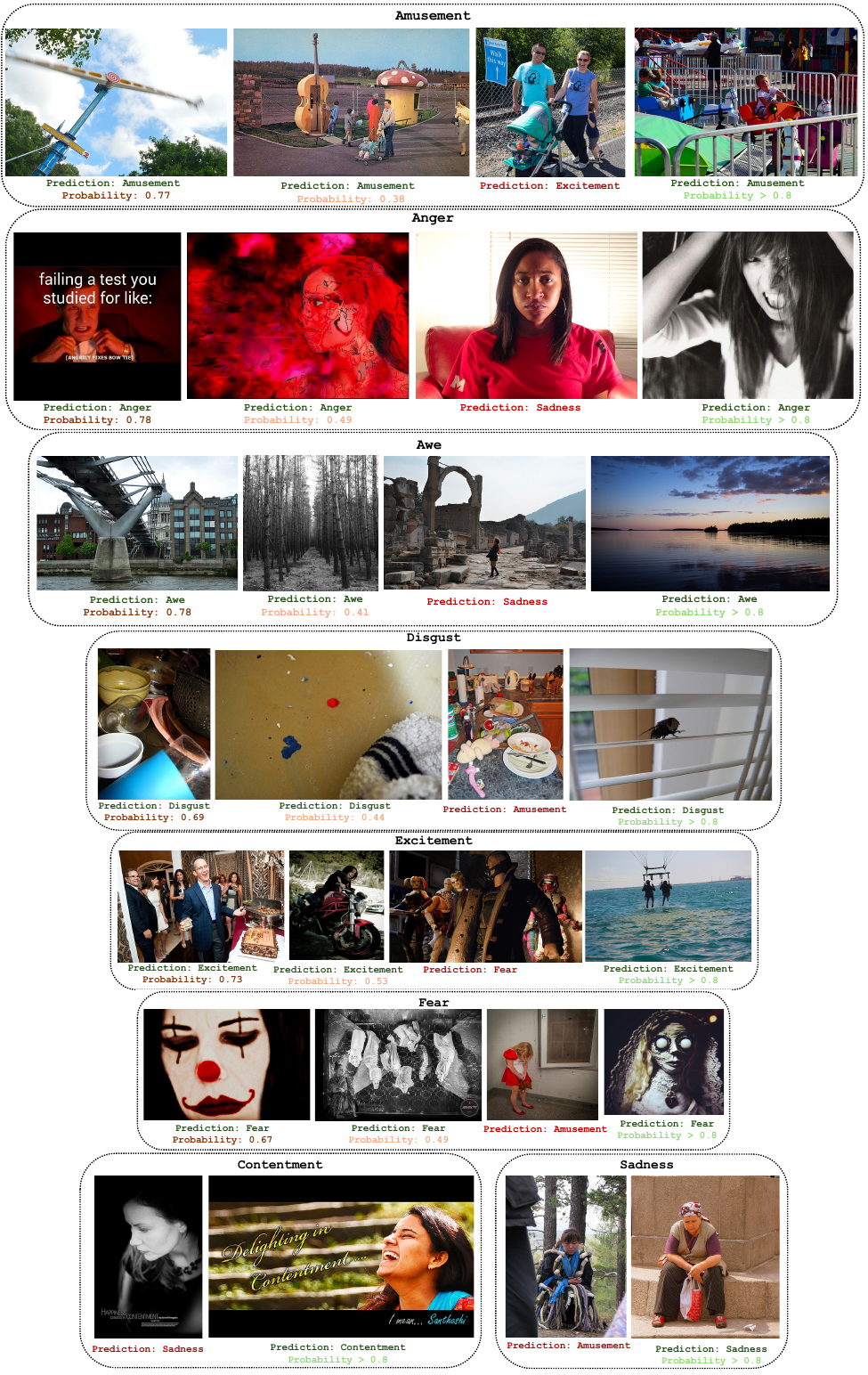}}
    \caption{Examples from the created \textbf{FI-Hard} dataset. Similar to EmoSet-Hard, for Contentment and Sadness, no instances are found that are predicted correctly with a probability less than 0.8. For all other categories, the two leftmost examples describe instances that are correctly predicted, but with a probability less than 0.8. The next example shows an image predicted incorrectly. Finally, the rightmost example for all categories show the correctly predicted samples, which have probability of prediction higher than 0.8.}
    \label{fig:benchmark_example_fi}
\end{figure*}

\subsection{Manual Analysis of Difficulty of Images}
\label{app:benchmark_difficulty}

We present examples from EmoSet-Hard and FI-Hard to demonstrate the qualitative difference in difficulty in predicting evoked emotions. Note that the \textit{examples may contain images that evoke strong negative emotions in the viewer}. As seen in Fig. \ref{fig:benchmark_example_emoset} and Fig. \ref{fig:benchmark_example_fi}, for each emotion category, the first 3 samples from the left are included in the final datasets, as they are either predicted correctly with a probability below 0.8 or are predicted incorrectly by the fine-tuned ViT models. 

Consider the examples from EmoSet-Hard described in Fig. \ref{fig:benchmark_example_emoset}. From the examples for \textit{Amusement}, the image with children is classified with the highest probability of belonging to this emotion class, followed by the image showing toys. The image of the squirrel, although predicted to belong to the Amusement class, is done so with a significantly lower probability. This hints at the bias within the dataset that leads models to associate certain elements in the image (children, toys, amusement parks, etc.) to the emotion class of Amusement. Thus, images with relatively uncommon elements, which may or may not be commonly associated with the Amusement emotion class, are included in the EmoSet-Hard set.
Another example of this can be seen in the images shown for \textit{Anger}, \textit{Disgust} and \textit{Fear} classes, where images that are more colorful or show toys or small children are classified into the Amusement category, disregarding the deeper context within the images. Further, as seen in the incorrectly classified example from the category of \textit{Awe}, the facial expressions of the children in the image lead the image to be misclassified to belong to Sadness. Thus, instances with relatively more uncommon elements are included in the EmoSet-Hard set based on our strategy. 

The examples from FI-Hard, as shown in Fig. \ref{fig:benchmark_example_fi} also testify to more difficult samples being chosen. The set includes images containing visual elements that can easily be correlated with certain emotion classes, but originally belong to different emotion categories. For instance, the misclassified images shown under \textit{Disgust} and \textit{Fear} categories contain toys, or colorfully dressed people. They are included in the FI-Hard dataset as potentially difficult instances to predict. 

\begin{figure*}
    \centering
    \includegraphics[width=0.8\textwidth]{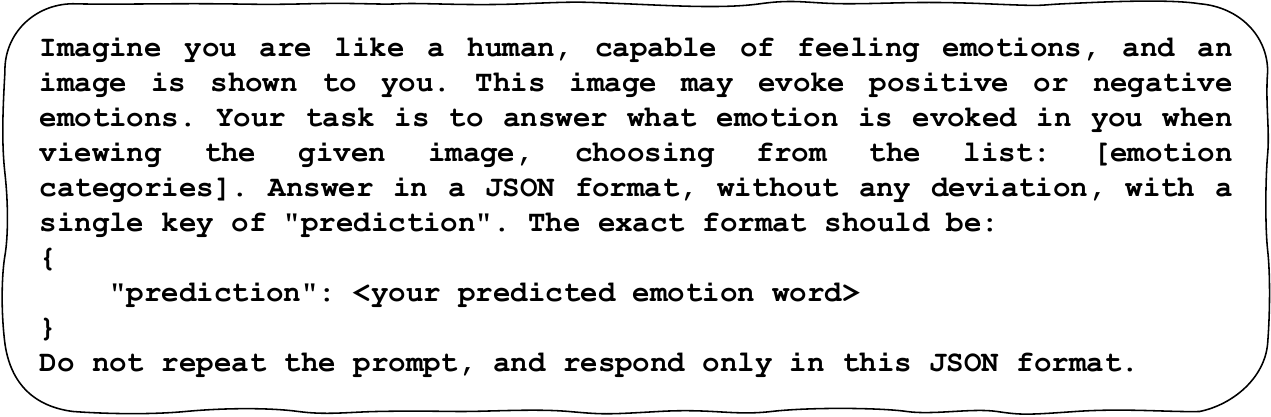}
    \caption{The prompt Simple Multimodal Classification}
    \label{fig:exp1_prompt}
\end{figure*}

\begin{figure*}
    \centering
    \includegraphics[width=0.7\textwidth]{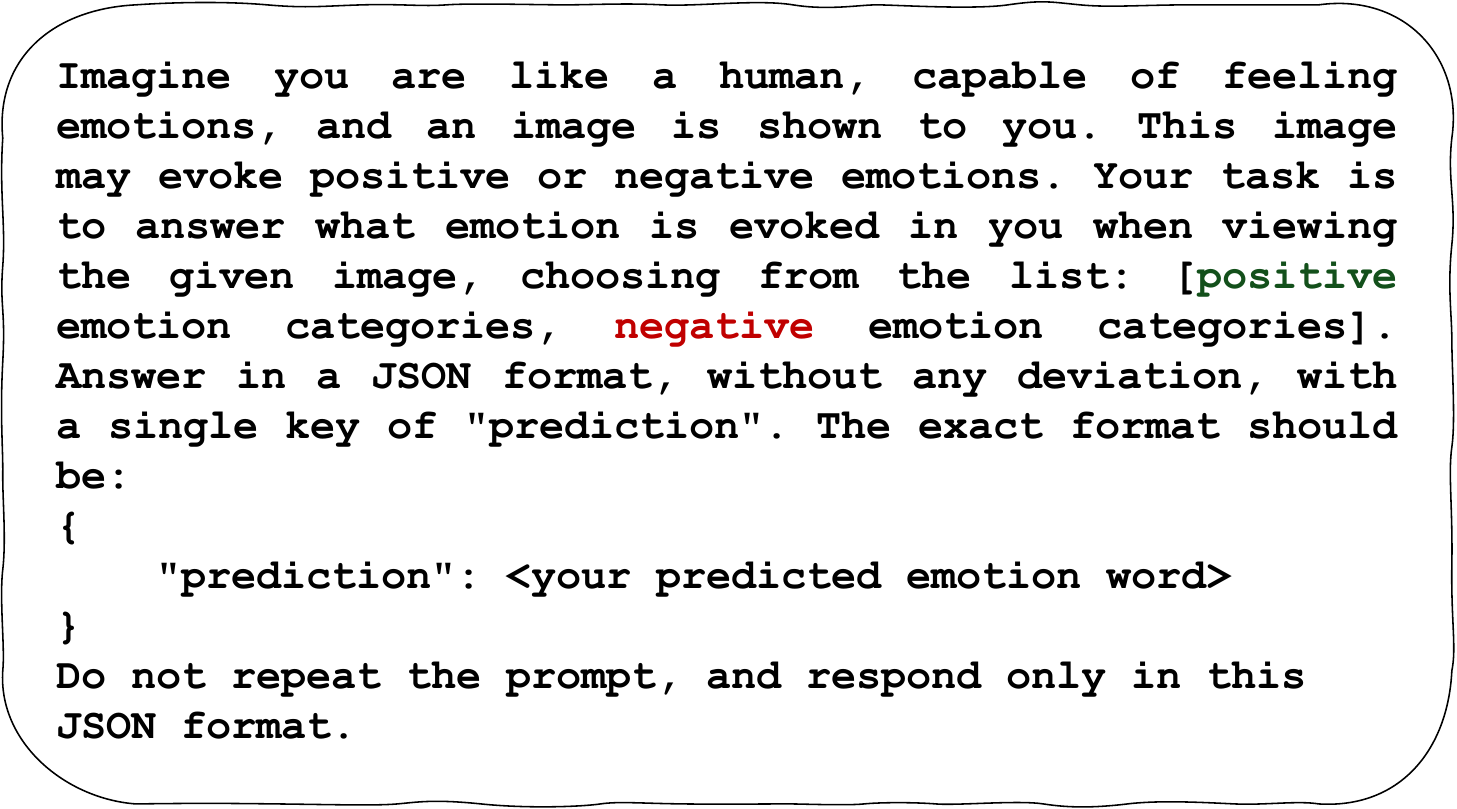}
    \caption{The prompt for shuffled order of emotions with positive emotions first.}
    \label{fig:shuffled_prompt_1}
\end{figure*}

\begin{figure*}
    \centering
    \includegraphics[width=0.7\textwidth]{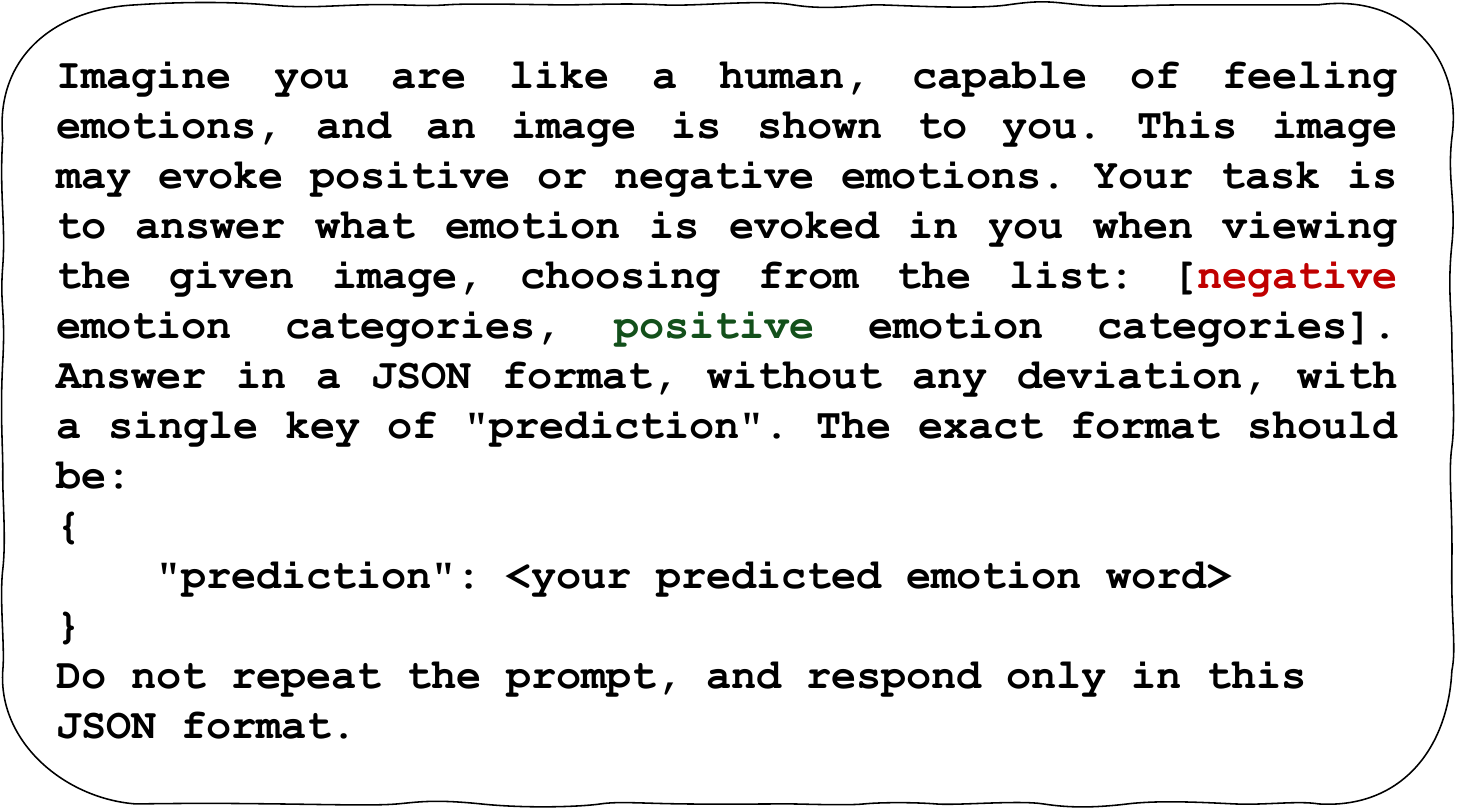}
    \caption{The prompt for shuffled order of emotions with negative emotions first.}
    \label{fig:shuffled_prompt_2}
\end{figure*}

\begin{figure*}
    \centering
    \includegraphics[width=0.7\textwidth]{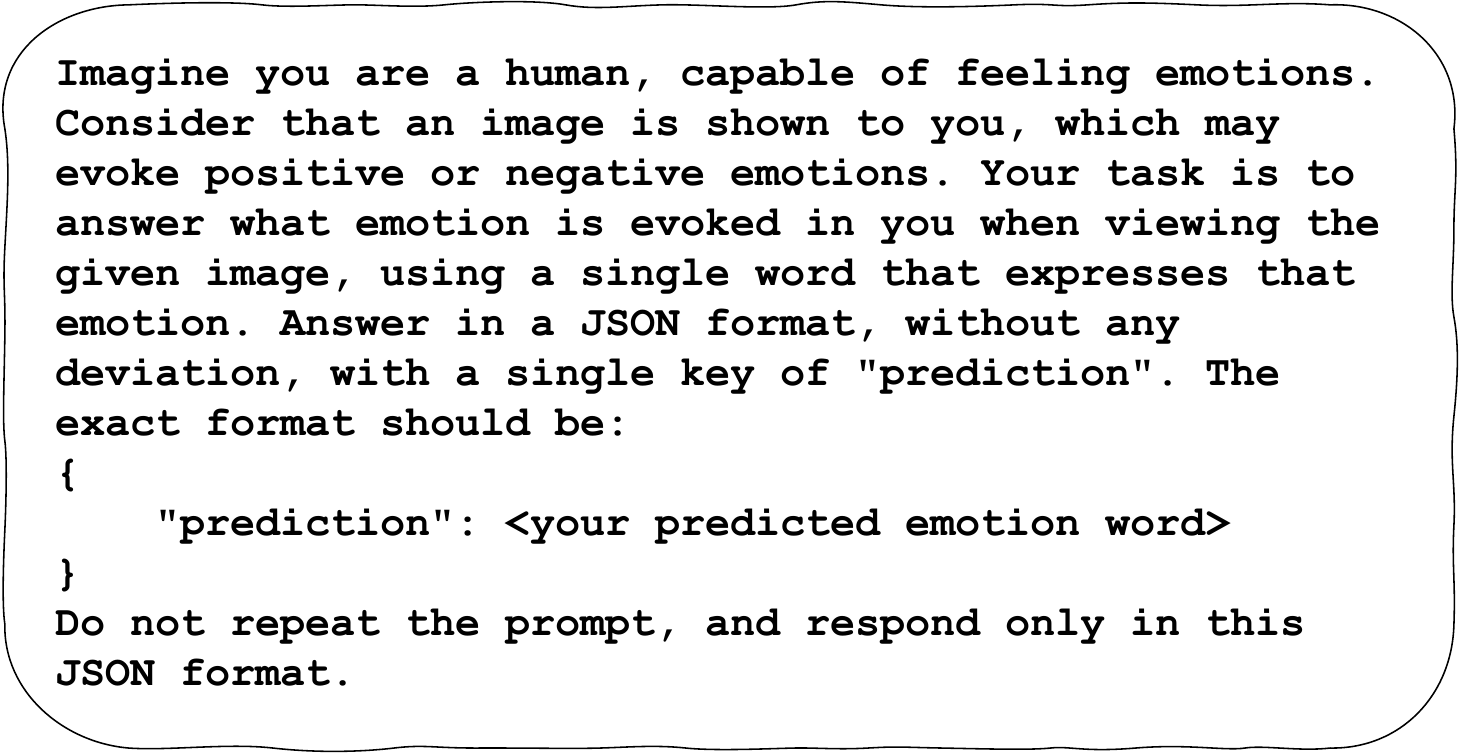}
    \caption{The prompt for open-vocabulary emotion prediction.}
    \label{fig:no_labeL_prompt}
\end{figure*}

\begin{figure*}
    \centering
    \includegraphics[width=0.7\textwidth]{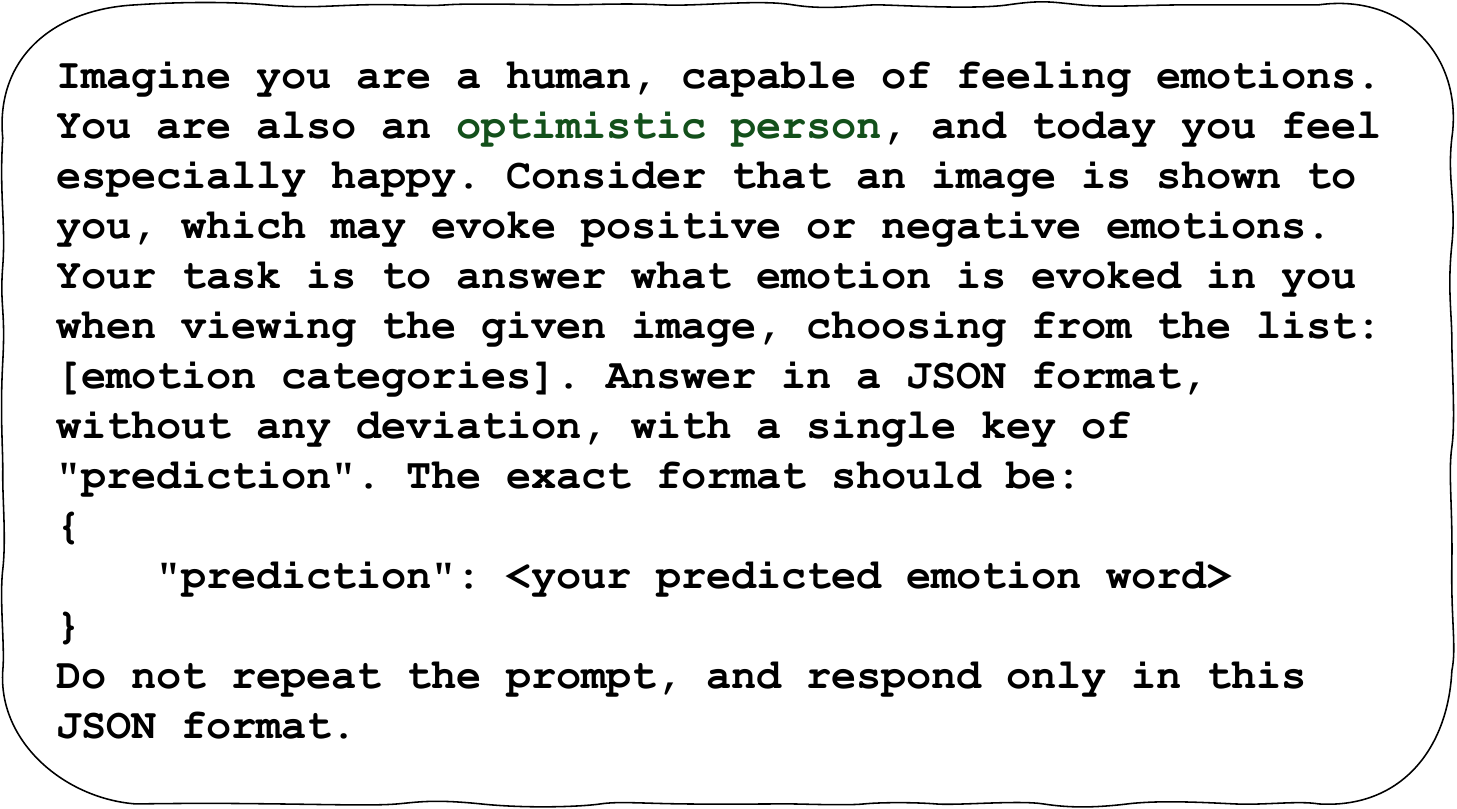}
    \caption{The prompt for adopting positive persona.}
    \label{fig:positive_persona_prompt}
\end{figure*}

\begin{figure*}
    \centering
    \includegraphics[width=0.7\textwidth]{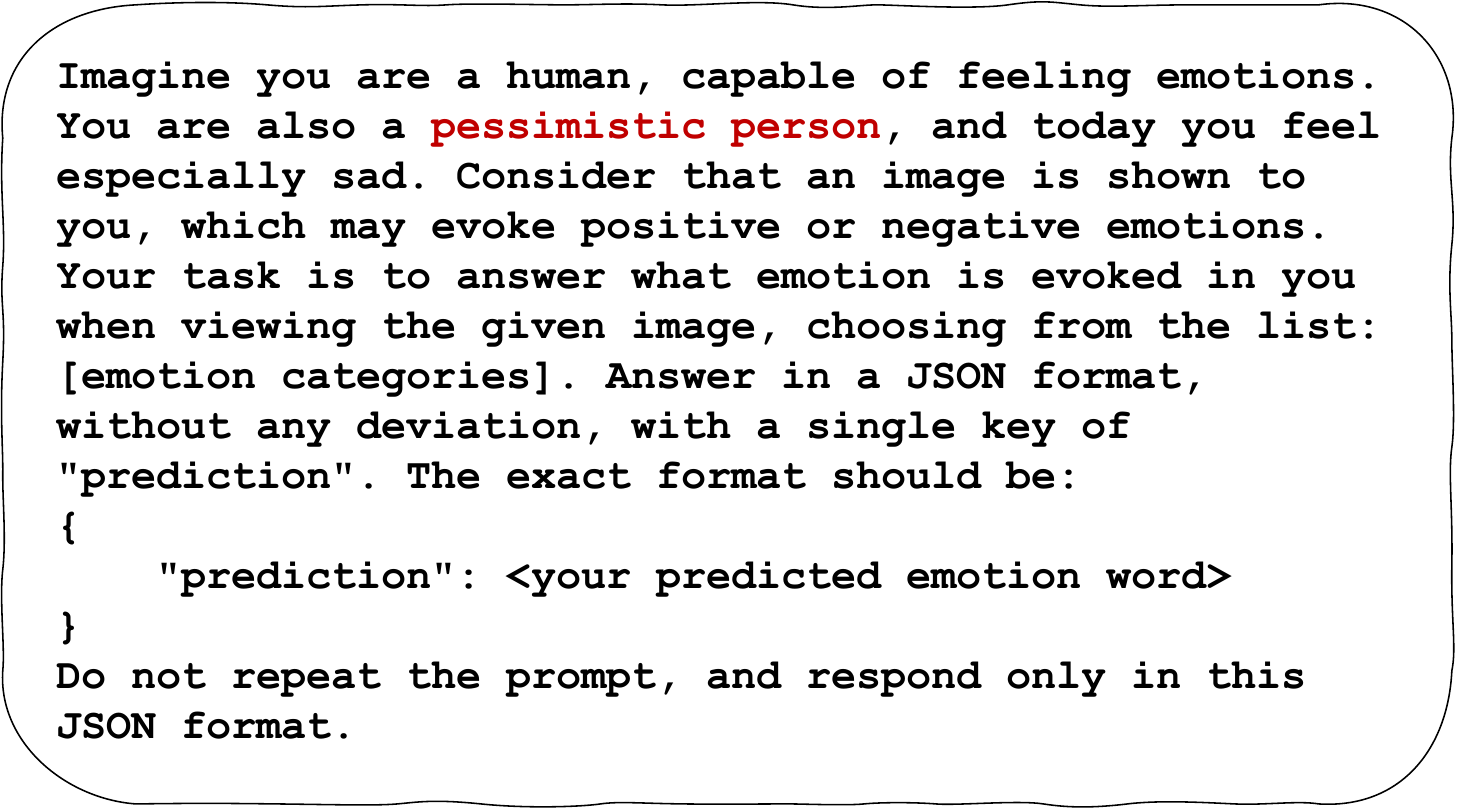}
    \caption{The prompt for adopting negative persona.}
    \label{fig:negative_persona_prompt}
\end{figure*}

\begin{figure*}
    \centering
    \includegraphics[width=0.7\textwidth]{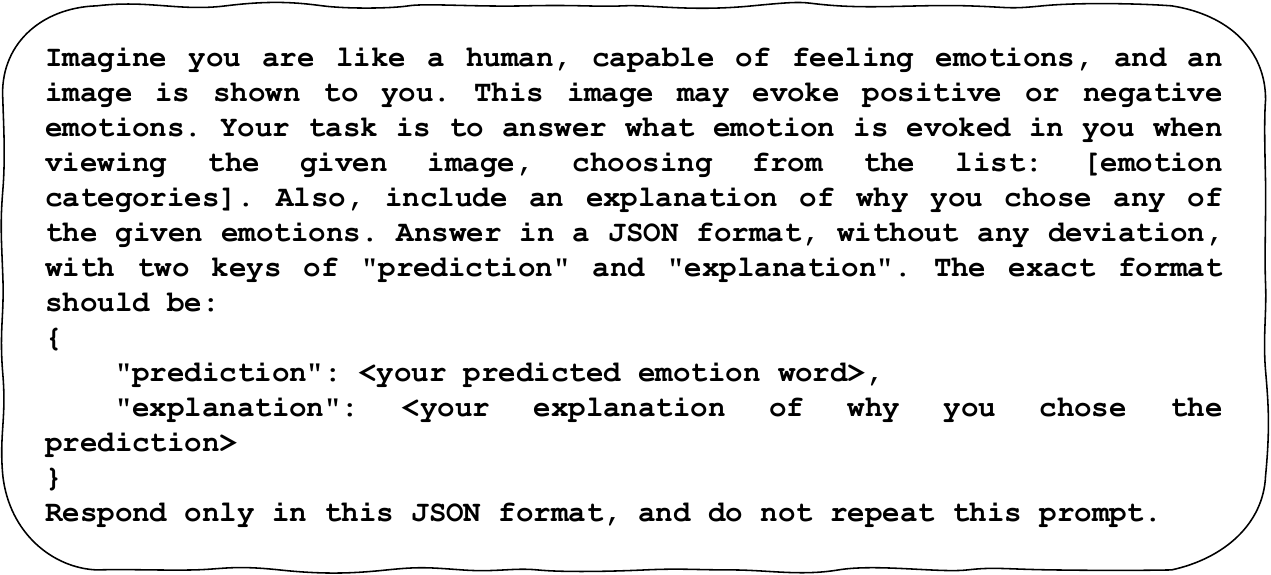}
    \caption{The prompt Explanation-based Reasoning}
    \label{fig:exp2_prompt}
\end{figure*}

\begin{figure*}
    \centering
    \includegraphics[width=0.7\textwidth]{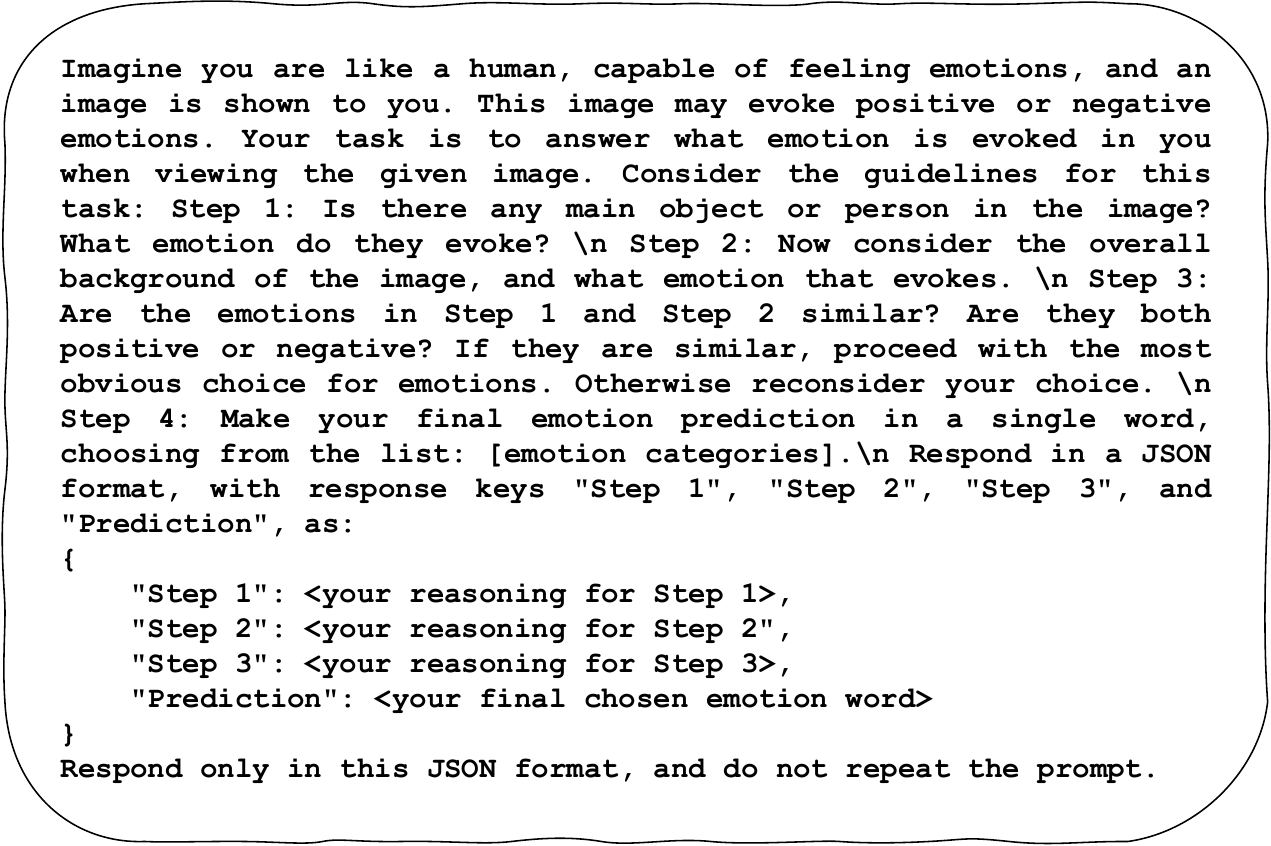}
    \caption{The prompt for Contextual Reasoning}
    \label{fig:exp3_prompt}
\end{figure*}

\begin{figure*}
    \centering
    \includegraphics[width=0.7\textwidth]{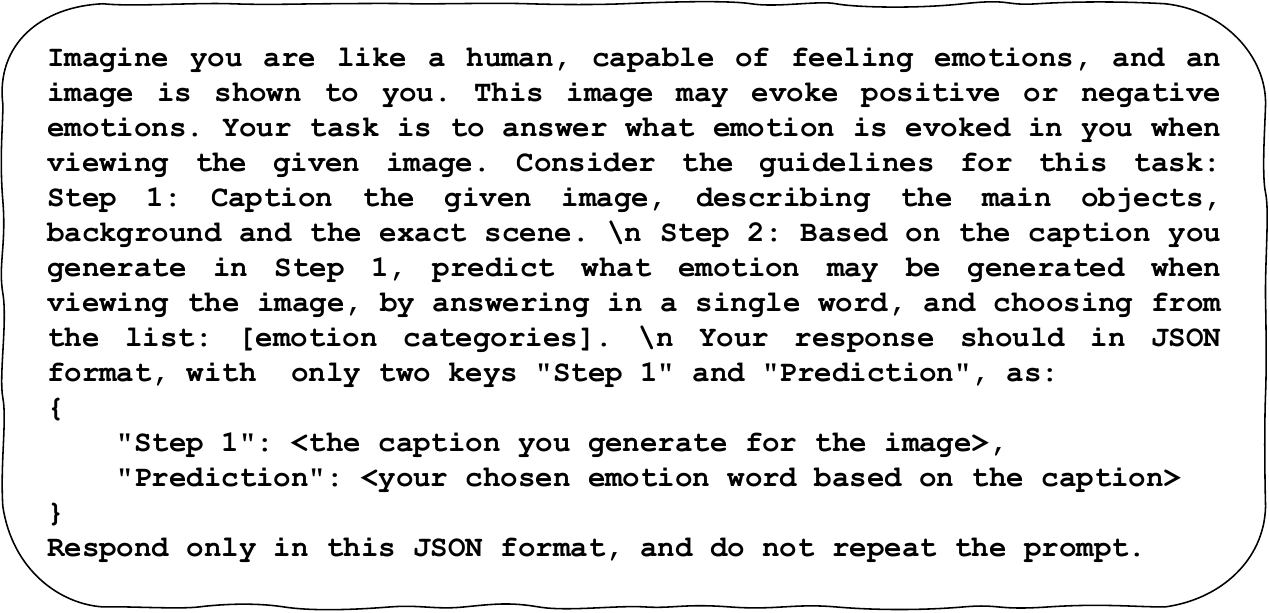}
    \caption{The prompt Caption-Based Reasoning}
    \label{fig:exp4_prompt}
\end{figure*}

\section{Main Experiments}

In this section, we provide additional details for all of our experiments. This includes details of implementation such as the resources, time or specific prompts used, and supplemental results.  

\subsection{Implementation Details}
\label{app:exp_impl}

To evaluate all of the open models, we use Huggingface \footnote{https://huggingface.co/models}. GPT4-o is evaluated using the OpenAI API \footnote{https://openai.com/index/openai-api/}. The open models are loaded in their full sizes, and run using GPUs (A40 with four cores). The maximum number of tokens to be generated is capped at 160, and is sufficient for all experiments. The time taken for the evaluation is influenced by the evaluation format, with the format of contextual reasoning (Section \ref{sec:reasoning}) taking the longest time, owing to the higher number of tokens required to be generated as output. The results reported are obtained through single runs of each type of prompt, owing to the significant computational and monetary resources required for using the models. 

\subsection{Emotion Properties Analyzed}
\label{app:emotion_properties}

We provide a categorization of the fine-grained emotion classes into broader positive and negative sentiment categories in Table \ref{tab:emotion_property_classification}. Note that we do this only for the emotion categories belonging to the popular 8-class emotion model \cite{mikels2005emotional}, as we consider only the constituent datasets adhering to this model of classification for the fine-grained analysis.
\begin{table}[t]
\small
    \centering
    \resizebox{\columnwidth}{!}{
    \begin{tabular}{ccc}
    \toprule
         Arousal/Sentiment & Positive & Negative \\
         \midrule
         High Arousal & Amusement, Excitement, & Fear, Anger, \\
         & Awe & Disgust \\
         Low Arousal & Contentment & Sadness \\
         \bottomrule
    \end{tabular}}
    \caption{Categorization of fine-grained emotion classes based on the broader Sentiment class and Arousal levels.}
    \label{tab:emotion_property_classification}
\end{table}

\subsection{Prompts Used}
\label{app:prompts}

We include the exact prompts included in this section, in the Figures \ref{fig:exp1_prompt} \ref{fig:shuffled_prompt_1}, \ref{fig:shuffled_prompt_2}, \ref{fig:no_labeL_prompt}, \ref{fig:positive_persona_prompt}, \ref{fig:negative_persona_prompt}, \ref{fig:exp2_prompt}, \ref{fig:exp3_prompt}, and \ref{fig:exp4_prompt}. 

We use a specific template format in the prompts, with the first line of each prompt being the following: "Imagine you are like a human, capable of feeling emotions, and an
image is shown to you.". We include this specifically to bypass content moderation policies in some models, that were otherwise leading the models to abstain from responding for some image samples. Although there was no overtly offensive or obscene content in the datasets we rely on, a large number of samples depict extreme (negative) emotions. We observed by experimenting with and without this specific starting line, that providing this warning helped in obtaining responses for most of the image samples. 

Also, in the current stage of our study, we include only zero-shot prompting strategies for evaluation. At the time of conducting experiments, some of the models included in the evaluation framework were incapable of reasoning over multiple visual inputs. Thus, providing other models with visual few-shot examples would give them an unfair edge. However, we do experiment with few-shot examples in textual form (results not included in this paper) for a small subset of the data. Precisely, we provide a caption-like description of images, along with the corresponding emotion evoked. We observe that this leads to further confused responses for models like LLaVA, and thus avoid using any few-shot examples for our large-scale evaluation experiments.

\subsection{Additional Results}
\label{app:additional_results}

\begin{figure}[t]
    \centering
    \includegraphics[width=\linewidth]{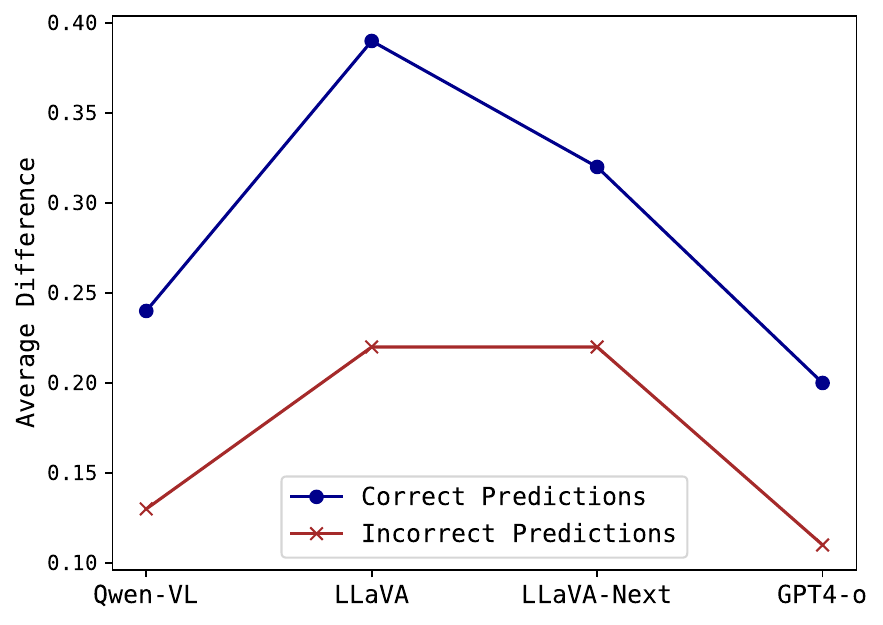}
    \caption{Average Difference between the most similar emotion class label and the next most similar emotion class label, given any model prediction, for both correct and incorrect predictions.}
    \label{fig:expected_diff_fine_grained}
\end{figure}

\begin{table*}[t]
    \centering
    \resizebox{\textwidth}{!}{
    \begin{tabular}{ccccccccc}
    \toprule
    Model Family & Amusement & Awe & Contentment & Excitement & Anger & Disgust & Fear & Sadness  \\
    \midrule
    GPT & 0.35 & 0.38 & \worst{0.31} & 0.49 & 0.36 & \best{0.63} & 0.54 & 0.40 \\
    LLaVA & 0.45 & 0.15 & 0.36 & 0.24 & \worst{0.09} & \best{0.47} & 0.21 & 0.37 \\
    LLaVA-Next & 0.44 & 0.24 & 0.31 & 0.37 & \worst{0.06} & \best{0.50} & 0.35 & 0.37 \\
    Qwen-VL & 0.29 & 0.27 & 0.25 & \best{0.52} & \worst{0.22} & 0.38 & 0.50 & 0.28\\
    \bottomrule
    \end{tabular}}
    \caption{Aggregated class-wise F1 scores for Simple Multimodal Classification. Model families include the F1 scores of each constituent model of different sizes (applicable for LLaVa and LLaVA-Next). The top-most F1 score achieved by each model family, across all fine-grained emotion classes, is highlighted in green, while the worst score is highlighted in red.}
    \label{tab:fine-grained-classwise-f1}
\end{table*}

\begin{table*}
    \centering
    \resizebox{\textwidth}{!}{
    \begin{tabular}{ccccc}
    \toprule
    Model Family & Positive-High Arousal & Positive-Low Arousal & Negative-High Arousal & Negative-Low Arousal \\
    \midrule
    GPT & 0.41 & \worst{0.31} & \best{0.51} & 0.40 \\ 
    LLaVA & 0.28 & 0.36 & \worst{0.26} & \best{0.37} \\ 
    LLaVA-Next & 0.35 & 0.31 & \worst{0.30} & \best{0.37} \\ 
    Qwen-VL & 0.36 & \worst{0.25} & \best{0.37} & 0.28 \\
    \bottomrule
    \end{tabular}}
    \caption{F1 scores for each (Sentiment, Arousal) category, averaged across model types, datasets, for the simple multimodal classification setting. The best and worst overall score for each model is highlighted in green and red respectively.}
    \label{tab:fine-grained-arousal}
\end{table*}

\begin{table*}
    \centering
    \resizebox{\textwidth}{!}{
    \begin{tabular}{ccccccccc}
    \toprule
    Model Family & Amusement & Awe & Contentment & Excitement & Anger & Disgust & Fear & Sadness  \\
    \midrule
    GPT & 0.41 & 0.30 & \worst{0.22} & 0.55 & \best{0.92} & 0.63 & 0.57 & 0.31 \\
    LLaVA & 0.42 & \worst{0.13} & 0.27 & 0.31 & \best{0.97} & 0.33 & 0.57 & 0.28 \\
    LLaVA-Next & 0.42 & 0.33 & \worst{0.23} & 0.75 & \best{0.95} & 0.43 & 0.50 & 0.28 \\
    Qwen-VL & 0.54 & 0.42 & \worst{0.15} & 0.41 & \best{0.85} & 0.42 & 0.58 & 0.45 \\
    \bottomrule
    \end{tabular}}
    \caption{Aggregated class-wise Precision scores for Simple Multimodal Classification. Model families include the Precision scores of each constituent model of different sizes (applicable for LLaVa and LLaVA-Next). The top-most Precision score achieved by each model family, across all fine-grained emotion classes, is highlighted in green, while the worst score is shown in red.}
    \label{tab:fine-grained-classwise-precision}
\end{table*}

\begin{table*}
    \centering
    \resizebox{\textwidth}{!}{
    \begin{tabular}{ccccccccc}
    \toprule
    Model Family & Amusement & Awe & Contentment & Excitement & Anger & Disgust & Fear & Sadness  \\
    \midrule
    GPT & 0.30 & 0.52 & 0.52 & 0.44 & \worst{0.23} & \best{0.62} & 0.51 & 0.55 \\
    LLaVA & 0.54 & 0.18 & 0.57 & 0.19 & \worst{0.05} & \best{0.78} & 0.15 & 0.60 \\
    LLaVA-Next & 0.48 & 0.38 & 0.50 & 0.26 & \worst{0.03} & \best{0.64} & 0.28 & 0.57 \\
    Qwen-VL & 0.20 & 0.20 & \best{0.75} & \best{0.75} & \worst{0.12} & 0.36 & 0.43 & 0.20 \\
    \bottomrule
    \end{tabular}}
    \caption{Aggregated class-wise Recall scores for Simple Multimodal Classification. Model families include the Recall scores of each constituent model of different sizes (applicable for LLaVa and LLaVA-Next). The top-most Recall score achieved by each model family, across all fine-grained emotion classes, is highlighted in green, while the worst scores are shown in red.}
    \label{tab:fine-grained-classwise-recall}
\end{table*}

\subsubsection{Fine-Grained Class-Wise Performance}

We present some additional results concerning the fine-grained performance of models. Table \ref{tab:fine-grained-classwise-f1} shows the average F1 scores achieved by each model family on the fine-grained emotion classes. The results are calculated by averaging scores on the EmoSet-Hard, FI-Hard, Abstract and ArtPhoto subsets of our benchmark, as they follow the 8-class classification of emotions. Interestingly, all model families, apart from Qwen-VL, consistently achieve the highest individual F1 score on the fine-grained category of disgust. Further, all models other than GPT4-o show the worst performance on the category of Anger. This is also in line with results presented in Section \ref{sec:exp_rq1}, where GPT4-o is seen to perform significantly better on negative emotion categories. In Table \ref{tab:fine-grained-arousal}, we also present an aggregate of the F1 scores, by grouping emotions based on the sentiment (positive or negative) and arousal (high or low) categories. 

We also show the class-wise Precision and Recall in Table \ref{tab:fine-grained-classwise-precision} and \ref{tab:fine-grained-classwise-recall}. It is worth noting that the precision scores for Anger are consistently the highest, while it is also the category with the worst F1 score for most models. In contrast to that, the recall scores for Anger are consistently the lowest, showing that a high number of false negatives affects the overall performance of models on this category the most. The recall scores are the highest for the Disgust category (except for Qwen-VL), which is also the class where models achieve the highest F1 scores. Overall, a complementary relationship can be seen for the precision and recall scores for most categories, and can be investigated deeply for further analyses and improvements in future work.

\subsubsection{Predicting Emotions Without Any Labels}
\label{app:no_label}

We study the capability of models to make fine-grained, distinct emotion predictions in further detail, in addition to the results presented in Section \ref{sec:no_label}. Recall that for all open-vocabulary prediction experiments, we calculate the semantic similarity of the model prediction with all emotion label classes, and assign the prediction to the class with the highest similarity. We now try to understand whether the model predicts an emotion that is truly closest semantically to a single emotion class, or it predicts a generic emotion word that could be considered almost as similar to multiple other emotion classes. In other words, we consider whether the maximum similarity score is significantly different from the second-largest similarity score between a given model prediction and the original emotion classes. Formally, given a model prediction \(o_i\), and the set of original class labels \(C\), we first calculate the maximum similarity to assign the prediction to a particular label class: 

\begin{equation}
\label{eq:maximum_similarity}
s_{\text{max}} = \max_{k}\, (\,sim\,(\,o_i, c_k\,))\, \forall\, c_k \in C
\end{equation}

Using this, we assign \(o_i\) to the label class as follows:

\begin{equation}
\label{eq:maximum_similarity_class}
j = \argmax_k\, (\,sim\,(\,o_i, c_k\,))\, \forall\, c_k \in C   
\end{equation}

\begin{equation}
\label{eq:prediction}
    \text{prediction}(o_i) = c_j
\end{equation}

Now, we calculate the second-largest similarity score as follows: 
\begin{equation}
\label{eq:second_maximum_similarity}
    s_{\text{second\_max}} = \max_{k}\, (\,sim\,(\,o_i, c_k\,))\, \forall\, c_k \in C \setminus c_j
\end{equation}
Then, we calculate the difference between the maximum possible similarity and the second maximum similarity between the model prediction and all the original label classes using Equations \ref{eq:maximum_similarity} and \ref{eq:second_maximum_similarity}, as: 

\begin{equation}
    {d} = s_{\text{max}} - s_{\text{second\_max}}
\end{equation}
Intuitively, \(d\) represents how different the final assigned class label is from the next most likely class label. We plot the expected values of \(d\) for all models in Fig. \ref{fig:expected_diff_fine_grained}. LLaVA (particularly LLaVA 13B) makes the most clearly distinguished predictions among all other models, as demonstrated by the highest expected difference between the maximum and second maximum similarity score for its predictions. Also, for all models, the difference is clearer when they make predictions that are eventually determined to belong to the correct label class. For incorrect predictions, the difference is much smaller, meaning that model predictions are unclear and have a close likelihood to belong to multiple different classes. GPT4-o makes the least well-distinguished predictions, which also agrees with the results presented in Fig. \ref{fig:no_label_f1} on the percentage of fine-grained predictions made by each model.

\subsubsection{Reasoning-Based Experiments}

We present additional results for the reasoning-based experiments. In Table \ref{tab:exp1_accuracies}, we first present the accuracy scores using the simple classification format where the corresponding model operation is \(M_{\text{c}}(\cdot)\). The best and worst-performing models are highlighted using green and red respectively. In Tables \ref{tab:exp2_accuracies}, \ref{tab:exp3_accuracies} and \ref{tab:exp4_accuracies}, we present the accuracy scores of models when prompted with explanation-based, contextual, and caption-based reasoning strategies respectively. For all of these tables, we use the green color to highlight the best-performing model for each dataset. Further, using orange, we designate the model that performed the worst for a particular dataset in the simplest setting (can be verified from Table \ref{tab:exp1_accuracies}) and accompany that with the changes due to the intervention applied. We show through this that the reasoning-based prompting strategies lead to improvements for most of the worst-performing initial combinations.

\begin{table*}
    \centering
    \small
    \begin{tabular}{cccccc}
        \toprule
         Model & Emotion6 & Abstract & ArtPhoto & FI & EmoSet (Hard) \\
         \midrule
         Qwen-VL & 54.2 & 26.5 & \colorbox{red!20}{35.7} & \colorbox{red!20}{31.3} & 42.5 \\
         \midrule
         LLaVA (7B) & \colorbox{red!20}{42.1} & \colorbox{green!20}{\textbf{29.1}} & 41.0 & \colorbox{green!20}{\textbf{59.2}} & \colorbox{red!20}{20.4} \\
         LLaVA (13B) & 59.6 & 20.1 & 42.8 & 38.8 & 34.7 \\
         \midrule
         LLaVA-NEXT (Vicuna 7B) & 57.9 & 27.3 & 40.7 & 56.8 & 26.2 \\
         LLaVA-NEXT (Mistral 7B) & 61.0 & \colorbox{red!40}{13.4} & 43.1 &  38.2 & 37.3 \\
         LLaVA-NEXT (Vicuna 13B) & 58.8 & 16.1 & 45.7 & 44.6 & 34.7 \\
         \midrule
         GPT4-o & \colorbox{green!40}{\textbf{66.4}} & 18.8 & \colorbox{green!20}{\textbf{48.4}} & 45.5 & \colorbox{green!20}{\textbf{45.6}} \\
         \midrule  
    \end{tabular}
    \caption{Average Accuracy Scores for Simple Multimodal Classification. The best and worst-performing models on each dataset
are highlighted in green and red colors respectively. The overall best-performing model is shown in a brighter
green color (GPT4-o on Emotion6), whereas the overall worst-performing model is shown in a brighter red color
(LLaVA-NEXT (Mistral 7B) on Abstract)}
    \label{tab:exp1_accuracies}
\end{table*}

\begin{table*}
    \centering
    \small
    \resizebox{\textwidth}{!}{
    \begin{tabular}{cccccc}
        \toprule
         Model & Emotion6 & Abstract & ArtPhoto & FI & EmoSet (Hard) \\
         \midrule
         Qwen-VL & 53.99 & \best{26.7} & \improved{39.87} \increase{(+4.11)} & \improved{27.94} \decrease{(-3.4)} & 45.26 \\
         \midrule
         LLaVA (7B) & \improved{49.5} \increase{(+7.3)} & 20.9 & 39.74 & 52.71 & \improved{19.6} \decrease{(-0.8)} \\
         LLaVA (13B) & 59.40 & 25.61 & 42.52 & 39.48 & 31.12 \\
         \midrule
         LLaVA-NEXT (Vicuna 7B) & 59.24 & 20.45  & 41.24 & \best{55.77} & 29.61 \\
         LLaVA-NEXT (Mistral 7B) & 61.45 & \improved{14.03} \increase{(+0.62)} & 42.74 & 40.31 & 35.91  \\
         LLaVA-NEXT (Vicuna 13B) & 55.73 & 20.71 & 45.43 \decrease & 43.68 & 35.66 \\
         \midrule
         GPT4-o & \best{66.11} & 17.72 & \best{48.83} & 43.98 & \best{45.45} \\
         \midrule  
    \end{tabular}}
    \caption{Accuracy Scores for Classification with Explanations. The changes in accuracy points (\%) compared to simple classification are shown alongside the actual values for the originally worst-performing models. The highest scores achieved are highlighted in green.}
    \label{tab:exp2_accuracies}
\end{table*}

\begin{table*}
    \centering
    \small
    \resizebox{\textwidth}{!}{
    \begin{tabular}{cccccc}
        \toprule
         Model & Emotion6 & Abstract & ArtPhoto & FI & EmoSet (Hard) \\
         \midrule
         Qwen-VL & 54.5 & 25.45 & \improved{34.2} \decrease{(-1.6)} & \best{44.4} \increase{(+13.05)} & 32.6 \\
         \midrule
         LLaVA (7B) & \improved{51.0} \increase{(+8.84)} & 18.4 & 27.9 & 36.77 & \improved{23.1} \increase{(+2.7)} \\
         LLaVA (13B) & 56.26  & 23.53 & 37.96 & 29.43 & 41.4  \\
         \midrule
         LLaVA-NEXT (Vicuna 7B) & 49.66 & 19.42 & 24.8 & 23.6 & 22.3 \\
         LLaVA-NEXT (Mistral 7B) & 59.5 & \improved{25.5} \increase{(+12.07)} & 36.9 & 33.0 & 46.17 \\
         LLaVA-NEXT (Vicuna 13B) & 57.48 & 23.21 & 38.5 & 39.15 & 34.6 \\
         \midrule
         GPT4-o & \best{64.72} & \best{30.35} & \best{49.57} & 41.0 & \best{48.34} \\
         \midrule  
    \end{tabular}}
    \caption{Accuracy Scores for Classification with Contextual Reasoning. The changes in accuracy points (\%) compared to simple classification are shown alongside the actual values for the originally worst-performing models. The highest scores achieved are highlighted in green.}
    \label{tab:exp3_accuracies}
\end{table*}

\begin{table*}
    \centering
    \small
    \resizebox{\textwidth}{!}{
    \begin{tabular}{cccccc}
        \toprule
         Model & Emotion6 & Abstract & ArtPhoto & FI & EmoSet (Hard) \\
         \midrule
         Qwen-VL & 56.67 & 17.51 & \improved{33.72} \decrease{(-2.04)} & \improved{31.56} \increase{(+0.2)} & 38.4 \\
         \midrule
         LLaVA (7B) & \improved{43.53} \increase{(+1.4)} & \best{25.9} & 38.43 & \best{44.04} & \improved{35.6} \increase{(+15.15)} \\
         LLaVA (13B) & 54.4 & 22.92 & 36.67 & 29.3 & 38.33 \\
         \midrule
         LLaVA-NEXT (Vicuna 7B) & 50.95 & 21.33 & 40.0 & 34.65 & 38.0  \\
         LLaVA-NEXT (Mistral 7B) & 60.58 & \improved{15.41} \increase{(+2.01)} & 43.71 & 33.57 & \best{47.41} \\
         LLaVA-NEXT (Vicuna 13B) & 55.5 & 23.04 & 39.16 & 31.67 & 40.69 \\
         \midrule
         GPT4-o & \best{65.82} & 20.77 & \best{48.82} & 40.65 & 46.21 \\
         \midrule  
    \end{tabular}}
    \caption{Accuracy Scores for Classification with Caption-Based Reasoning. The changes in accuracy points (\%) compared to simple classification are shown alongside the actual values for the originally worst-performing models. The highest scores achieved are highlighted in green.}
    \label{tab:exp4_accuracies}
\end{table*}

\begin{figure}[t]
    \centering
    \includegraphics[width=\linewidth]{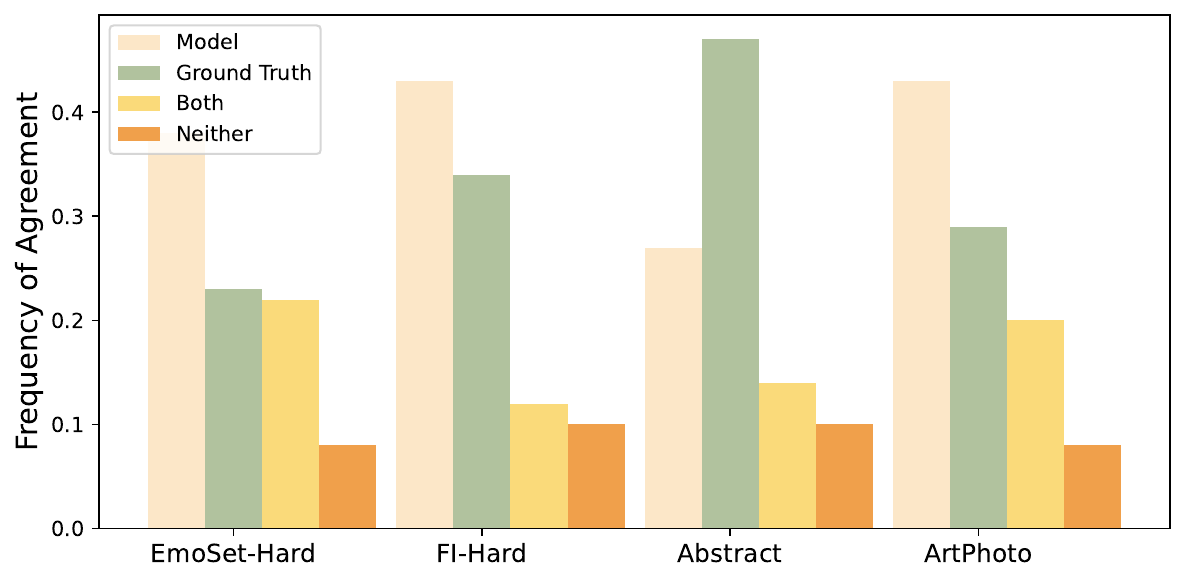}
    \caption{The frequency of agreement with model predicted label (cream/beige), the dataset ground truth label (green), both labels (yellow), and neither label (orange), for each dataset, averaged across all models studied.}
    \label{fig:humeval_datasets}
\end{figure}

\subsubsection{Error Analysis and Human Evaluation}
\label{app:humeval}

We provide additional details about the analysis of model errors and the manual evaluation conducted. To create the different error categories, as we characterize specific emotion classes based on their sentiment and arousal, we focus only on EmoSet-Hard, FI-Hard, Abstract and ArtPhoto, as they use the same 8-class labels. Based on the 3 different categories of errors defined in Section \ref{sec:exp_rq3}, we sample around 10 error examples for each category, for each model and dataset. It leads to a total of around 500 error samples being annotated. We consider only the smallest variants for LLaVA (7B) and LLaVA-Next (Vicuna 7B) for the current stage of the study. Our annotators are primarily graduate student volunteers from the authors' team. We specifically use annotations from people aware of research in computing for emotions, owing to several reasons: (a) the current stage of the study is small-scale, (b) clear knowledge about evoked emotions and nuanced emotion categories is valuable for our study, and (c) to ensure high-quality annotations, which is usually compromised when aggregating large numbers of crowd-sourced annotations. The annotators are located geographically within North America. For each annotation turn, the visual stimuli is displayed for about 3-5 seconds, along with the model prediction and the dataset ground truth. However, it is not disclosed which label is model prediction and which one is the ground truth to eliminate any bias in annotation. 

In addition to the results we present in Section \ref{sec:exp_rq3}, we also include an additional analysis of how human agreement varies with each dataset considered. We plot the average frequency (across all models) of human agreement with the model predictions, ground truth label or both, for each dataset in Fig. \ref{fig:humeval_datasets}. For all datasets, on average, human annotations agree with the model predictions more often. This is not the case only for Abstract, where human agreement with the dataset ground truth is significantly higher than with model predictions, meaning that Abstract provides the most reliable ground truth labels. 

\end{document}